\documentclass{article}
\PassOptionsToPackage{numbers, compress}{natbib}
\usepackage[final]{neurips_2021}
\usepackage[utf8]{inputenc}
\usepackage[T1]{fontenc}
\usepackage[noend]{algpseudocode}
\usepackage{algorithm}
\usepackage{amsfonts}
\usepackage{amsmath,amssymb}
\usepackage{booktabs}
\usepackage{graphicx}
\usepackage{hyperref}
\usepackage{macros}
\usepackage{microtype}
\usepackage{nicefrac}
\usepackage{subcaption}
\usepackage{tabularx}
\newcolumntype{C}{>{\centering\arraybackslash}X}
\usepackage{url}
\usepackage{xcolor}
\usepackage{wrapfig}
\newif\ifdark
\IfFileExists{/Users/vedaldi/.bash_profile}{
\makeatletter
\immediate\write18{\unexpanded{%
if defaults read -g AppleInterfaceStyle 2>/dev/null;
then echo \\darktrue > /tmp/displaymode.tex;
else echo \\darkfalse > /tmp/displaymode.tex; fi}}
\makeatother 
\input{/tmp/displaymode.tex}
\ifdark
\definecolor{pcolor}{HTML}{1E1E1E}
\definecolor{tcolor}{HTML}{C5C5C5}
\else
\definecolor{pcolor}{HTML}{FDF6E3}
\definecolor{tcolor}{HTML}{333333}
\fi
\pagecolor{pcolor}
\color{tcolor}
\hbadness=\maxdimen
\vbadness=\maxdimen
\vfuzz=30pt
\hfuzz=30pt
}{}
\def\x{$\times$}

\usepackage{pifont}% http://ctan.org/pkg/pifont
\newcommand{\xmark}{\ding{55}}%

\makeatletter
\renewcommand{\paragraph}{%
  \@startsection{paragraph}{4}%
  {\z@}{0em}{-1em}%
  {\normalfont\normalsize\bfseries}%
}
\newcommand{\midsepadjust}{\aboverulesep = 0.305mm \belowrulesep = 0.684mm}
\midsepadjust

\makeatother

\title{Keeping Your Eye on the Ball: \\ Trajectory Attention in Video Transformers}

\author{%
  Mandela Patrick\thanks{Equal contribution.} \\
  Facebook AI \\
  \texttt{mandelapatrick@fb.com} \\
  \And
   Dylan Campbell\footnotemark[1] \\
   University of Oxford \\
   \texttt{dylan@robots.ox.ac.uk} \\
   \And
   Yuki Asano\footnotemark[1] \\
   University of Oxford \\
   \texttt{yuki@robots.ox.ac.uk} \\
   \AND
   Ishan Misra \\
   Facebook AI \\
   \texttt{imisra@fb.com} \\
   \And
   Florian Metze \\
   Facebook AI \\
   \texttt{fmetze@fb.com} \\
   \And
   Christoph Feichtenhofer \\
   Facebook AI \\
   \texttt{feichtenhofer@fb.com} \\
   \AND
   Andrea Vedaldi \\
   Facebook AI \\
   \texttt{vedaldi@fb.com} \\
   \And
   Jo\~{a}o F. Henriques \\
   University of Oxford \\
   \texttt{joao@robots.ox.ac.uk} \\
}

\begin{document}
\maketitle
\begin{abstract}
In video transformers, the time dimension is often treated in the same way as the two spatial dimensions.
However, in a scene where objects or the camera may move, a physical point imaged at one location in frame $t$ may be entirely unrelated to what is found at that location in frame $t+k$.
These temporal correspondences should be modeled to facilitate learning about dynamic scenes.
To this end, we propose a new drop-in block for video transformers---\emph{trajectory attention}---that aggregates information along implicitly determined motion paths.
We additionally propose a new method to address the quadratic dependence of computation and memory on the input size, which is particularly important for high resolution or long videos.
While these ideas are useful in a range of settings, we apply them to the specific task of video action recognition with a transformer model and obtain state-of-the-art results on the Kinetics, Something--Something V2, and Epic-Kitchens datasets.  
Code and models are available at: \url{https://github.com/facebookresearch/Motionformer}.  
% The adoption of the transformer architecture has led to a new paradigm for both natural language and image understanding tasks.
% Recently it has been shown to easily extend to video data, where the additional dimension of time leads to an increased complexity.
% However, simply extending image-based approaches to video data is not enough.
% Instead, more explicit modelling of time can yield substantial improvements.
% In this paper, we propose an approach that boosts its performance by leveraging additional motion information inherent to the hidden layers of transformers and without the explicit calculation of optical flow.
% This additional signal about temporal aspects models image patches' trajectory through time and we show that this new drop-in block for video transformers---\emph{trajectory attention}---substantially outperforms the previous state-of-the-art video models for action recognition.
% We additionally propose a new method to address the quadratic dependence of computation/memory on input size, which becomes even more important for long videos.
% Combining our two contributions, we set state-of-the-art video action recognition results on the following 4 datasets: Kinetics-400, Kinetics-600, Something--Something V2, and Epic-Kitchens.
\end{abstract}

\section{Introduction}\label{s:intro}

Transformers~\cite{vaswani17attention} have become a popular architecture across NLP~\cite{bert18_naccl}, vision~\cite{dosovitskiy2021an} and speech~\cite{wav2vec}.
The self-attention mechanism in the transformer works well for different types of data and across domains.
However, its generic nature and its lack of inductive biases also mean that transformers typically require extremely large amounts of data for training~\cite{radford2019language, gpt3}, or aggressive domain-specific augmentations~\cite{touvron2020training}.
This is particularly true for video data, for which transformers are also applicable~\cite{neimark2021video}, but where statistical inefficiencies are exacerbated.
While videos carry rich temporal information, they can also contain redundant spatial information from neighboring frames.
Vanilla self-attention applied to videos compares pairs of image patches extracted at all possible spatial locations and frames.
This can lead it to focus on the redundant spatial information rather than the temporal information, as we show by comparing normalization strategies in our experiments.
% As our experiments show, this fails to discover the temporal information as it focuses on the redundant spatial information, to the point that randomly reordering the frames has little effect on the performance.
% We do show that this is the case in Table 4 where there is a big drop between normalizing across space, and normalizing across space-time; but this is an ablation of our attn, so is not strictly comparable to vanilla attn.

We therefore contribute a variant of self-attention, called \emph{trajectory attention}, which is better able to characterize the temporal information contained in videos.
For the analysis of still images, spatial locality is perhaps the most important inductive bias, motivating the design of convolutional networks~\cite{lecun1998gradient} and the use of spatial encodings in vision transformers~\cite{dosovitskiy2021an}.
This is a direct consequence of the local structure of the physical world:
points that belong to the same 3D object tend to project to pixels that are close to each other in the image.
By studying the correlation of nearby pixels, we can thus learn about the objects.

\begin{figure}[t]
    \centering
    %\vspace{-5mm}
    \includegraphics[width=0.9\linewidth]{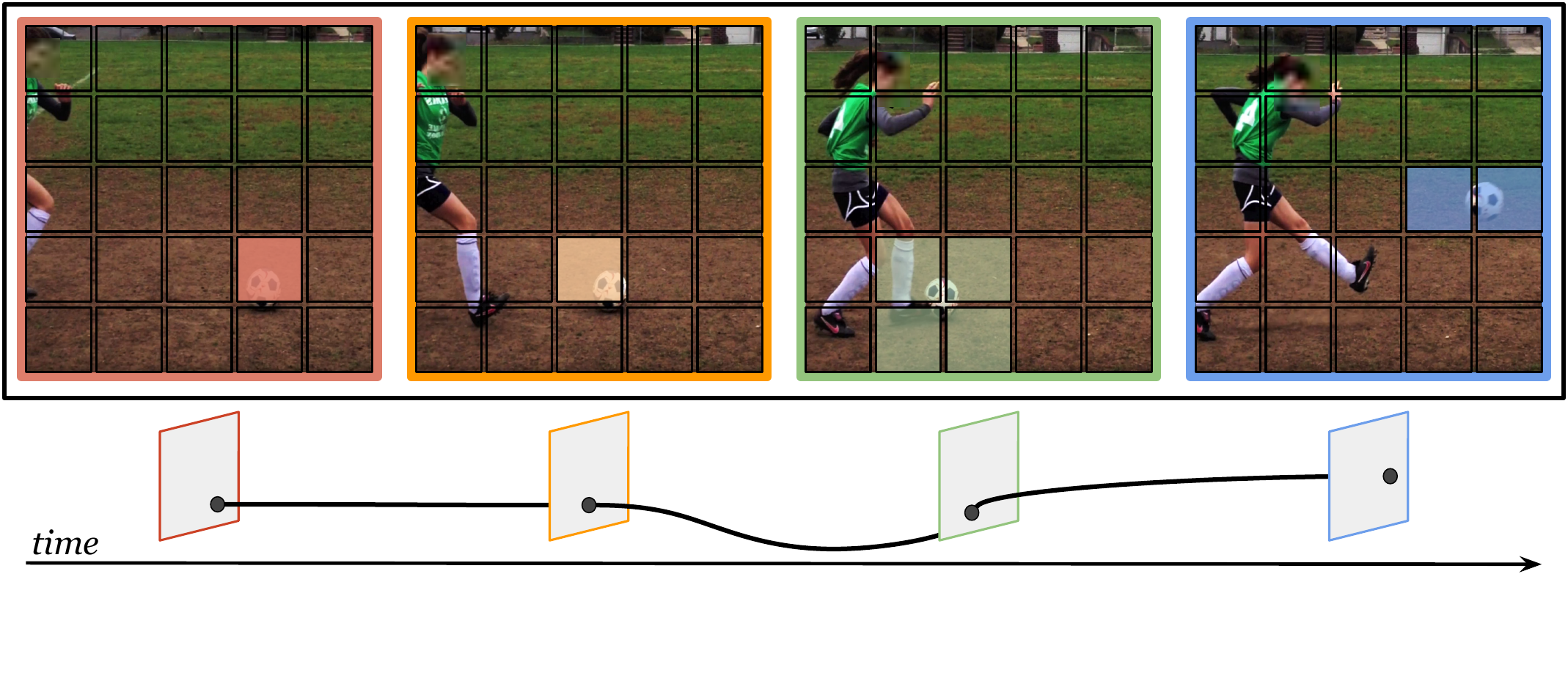}
    % \vspace{-5mm} 
    \caption{
        \textbf{Trajectory attention.}
        In this sequence of frames from the Kinetics-400 dataset, depicting the action `kicking soccer ball', the ball does not remain stationary with respect to the camera, but instead moves to different locations in each frame.
        Trajectory attention aims to share information along the motion path of the ball, a more natural inductive bias for video data than pooling axially along the temporal dimension or over the entire space-time feature volume.
        This allows the network to aggregate information from multiple views of the ball, to reason about its motion characteristics, and to be less sensitive to camera motion.
    }\label{fig:teaser}
    \vspace{-3mm}
\end{figure}

% \begin{figure}[t]
% \begin{center}
% \includegraphics[width=0.99\linewidth]{figs/vid_1.png}
% \vspace{-1em}
% \end{center}
%   \caption{\textbf{Joint Space-Time vs Space-Time Motion Attention.}
%   Here we show that Space-Time Motion basically tracks objects across time due to our normalization trick, modelling motion and displacements in time.  
%   \label{fig:splash}
%   }
% \end{figure}

% \begin{figure}
%     \centering
%     \includegraphics[width=0.99\linewidth]{figs/splash_motention-2.pdf}
%     \caption{
% Pair of adjacent frames from Kinetics-400 showing the action `kicking soccer ball'. First frame: ball and boot inside a single patch. Second frame: ball and boot in separate patches after the kick. Motion path arrows between the two. We want to show that the aggregated motion feature sums both patch features to share information with the single patch in the first frame. Also display the pixel flow of point on ball in contrast (single motion arrow). Also include a patch in the second frame with another boot, to demonstrate the possibility of incorrect matches due to ambiguities in the scene.
% }
%     \label{fig:patchflow}
% \end{figure}

%\vspace{-5mm}

Videos are similar, except that 3D points \emph{move} over time, thus projecting on different parts of the image along certain 2D \emph{trajectories}.
Existing video transformer methods~\cite{bertasius2021spacetime, arnab2021vivit, neimark2021video} disregard these trajectories, pooling information over the entire 3D space-time feature volume~\cite{arnab2021vivit, neimark2021video}, or pooling axially across the temporal dimension~\cite{bertasius2021spacetime}.
We contend that pooling along motion trajectories would provide a more natural inductive bias for video data,
allowing the network to aggregate information from multiple views of the same object or region, to reason about how the object or region is moving (for example, the linear and angular velocities), and to be invariant to camera motion.

We leverage attention itself as  a mechanism to find these trajectories.
This is inspired by methods such as RAFT~\cite{teed20raft:}, which showed that excellent estimates of optical flow can be obtained from the correlation volume obtained by comparing local features across space and time.
We observe that the joint attention mechanism for video transformers computes such a correlation volume as an intermediate result.
However, subsequent processing collapses the volume without consideration for its particular structure.
In this work, we seek instead to use the correlation volume to guide the network to pool information along motion paths.

We also note that visual transformers operate on image patches which, differently from individual pixels, cannot be assumed to correspond to individual 3D points and thus to move along simple 1D trajectories.
For example, in \figref{fig:teaser}, depicting the action `kicking soccer ball', the ball spans up to four patches, depending on the specific video frame.
Furthermore, these patches contain a mix of foreground (the ball) and background objects, thus at least two distinct motions.
Fortunately, we are not forced to select a single putative motion: the attention mechanism allows us to assemble a motion feature from all relevant `ball regions'.

Inspired by Nystr\"omformer~\cite{xiong2021nystromformer}, we also propose a principled approximation to self-attention, \emph{Orthoformer}.
Our approximation sets state-of-the-art performance on the recent Long Range Arena (LRA) benchmark~\cite{tay2021long} for evaluating efficient attention approximations and generalizes beyond the video domain to long text and high resolution images, with lower FLOPS and memory requirements compared to alternatives, Nystr\"omformer and Performer~\cite{choromanski2021rethinking}.
Combining our approximation with trajectory attention allows us to
% efficiently use large video sequences and achieve \textcolor{red}{XXX}. % Experiments too late...
significantly improve its computational and memory efficiency.
With our contributions, we set state-of-the-art results on four video action recognition benchmarks.

\section{Related Work}

\paragraph{Video representations and 3D-CNNs.}
Hand-crafted features were originally used to convert video data into a representation amenable to analysis by a shallow linear model. 
% Such representations include SIFT-3D~\cite{Scovanner07}, HOG3D~\cite{klaser2008spatio}, Motion Boundary Histogram~\cite{dalal2006human}, Cuboids~\cite{dollar2005behavior}, Action Bank~\cite{sadanand2012action} and IDT~\cite{wang2013action}.
Such representations include SIFT-3D~\cite{Scovanner07}, HOG3D~\cite{klaser2008spatio}, and IDT~\cite{wang2013action}.
Since the breakthrough of AlexNet~\cite{Krizhevsky12} on the ImageNet classification benchmark~\cite{imagenet}, which demonstrated the empirical benefits of deep neural networks to learn representations end-to-end, there have been many attempts to do the same for video.
Architectures with 3D convolutions---3D-CNNs---were originally proposed to learn deep video representations~\cite{Tran15}. 
Since then, improvements to this paradigm include the use of ImageNet-inflated weights~\cite{I3D}, the space-time decomposition of 3D convolutions~\cite{P3D, tran2018closer, xie2018rethinking}, channel-separated convolutions~\cite{Tran_2019}, non-local blocks~\cite{XiaolongWang18}, and attention layers~\cite{doubleattention}. 
Optical flow-based pooling can be used instead of temporal convolutions to improve the representation's robustness to camera and object motions \cite{afouras2020self}. 
Our approach shares this motivation.
% \cite{afouras2020self} pre-trained optical flow to average attention over trajectories, which has a similar inspiration

\paragraph{Vision transformers.} The transformer architecture~\cite{vaswani17attention}, originally proposed for natural language processing, has recently gained traction in the computer vision domain. 
The vision transformer (ViT)~\cite{dosovitskiy2021an} decomposes an image into a sequence of $16\times16$ words and uses a multi-layer transformer to perform image classification.
%While demonstrating state-of-the-art performance when trained on large-scale datasets such as JFT300M~\cite{sun17JFT} and IM-21k~\cite{imagenet}, it performs less convincingly on smaller datasets.
To improve ViT's data efficiency, DeiT~\cite{touvron2020training} used distillation from a strong teacher model and aggressive data augmentation.
Transformers have also been used in a variety of vision image tasks, such as image representation learning~\cite{chen2020generative, wu2020visual, desai2021virtex, sariyildiz2020learning}, image generation~\cite{parmar2018image}, object detection~\cite{locatello2020objectcentric, carion2020endtoend}, video question-answering~\cite{kant2020spatially}, few-shot learning~\cite{doersch2020crosstransformers}, and image--text~\cite{lu2019vilbert, su2019vlbert, Tan_2019, li2019visualbert, tan2020vokenization}, video-text~\cite{sun2019videobert, sun2019contrastive_arxiv, zhu2020actbert, gabeur2020multimodal, patrick2020supportset, akbari2021vatt, bain2021frozen}, and video-audio~\cite{lee2021parameter, patrick2021spacetime, huang2021multilingual} representation learning. 
% While the use of transformer architectures for video is still in its infancy, concurrent works~\cite{bertasius2021spacetime, arnab2021vivit, neimark2021video, fan2021multiscale} have already demonstrated that this is a highly promising direction.
% % video action recognition~\cite{XiaolongWang18, girdhar2019video,timesformer}, video question-answering~\cite{kant2020spatially},
% However, these approaches do not have a mechanism for reasoning about motion paths, treating time as just another dimension, unlike our approach.
% and instead pool information over the entire space-time volume or axially along the temporal dimension.

% video action recognition~\cite{XiaolongWang18, girdhar2019video,timesformer}, 

\paragraph{Attention for video recognition.} The self-attention operation proposed in the transformer~\cite{vaswani17attention} have been adapted to video recognition tasks.
Wang et al.~\cite{XiaolongWang18} propose the non-local mean operation for video action recognition, which is equivalent to the standard transformer self-attention applied uniformly across space and time, while our proposed trajectory attention does not treat the space and time dimensions equivalently.
Zhao et al.~\cite{zhao2018trajectory} propose a CNN architecture that explicitly predicts trajectories and aggregates information along them using a convolution operation. 
In contrast, our transformer architecture does not explicitly predict trajectories, but instead provides an inductive bias that encourages the network to consider motion trajectories where useful.
Concurrent works~\cite{bertasius2021spacetime, arnab2021vivit, neimark2021video, fan2021multiscale} have also adapted the self-attention operation to the spatio-temporal nature of videos, however, these approaches do not have a mechanism for reasoning about motion paths, treating time as just another dimension, unlike our approach.

\paragraph{Efficient attention.}
Due to the quadratic complexity of self-attention, there has been a significant amount of research on how to reduce its computational complexity with respect to time and memory use.
Sparse attention mechanisms~\cite{child2019sparsetransformer} were used to reduce self-attention complexity to $\cO(n\sqrt{n})$, and locality-sensitivity hashing was used by Reformer~\cite{Kitaev2020Reformer} to further reduce this to $\cO(n\log n)$.
More recently, linear attention mechanisms have been introduced, namely Longformer~\cite{Beltagy2020Longformer}, Linformer~\cite{wang2020linformer}, Performer~\cite{choromanski2021rethinking} and Nystr\"omformer~\cite{xiong2021nystromformer}. 
The Long Range Arena benchmark~\cite{tay2021long} was recently introduced to compare these different attention mechanisms.

\paragraph{Temporal correspondences and optical flow.} 
% Why are we mentioning this? These are explicit ways to establish correspondences, which we use as inspiration for our implicit model
There are many approaches that aim to establish explicit correspondences between video frames as a way to reason about camera and object motion.
For short-range correspondences across time, optical flow algorithms~\cite{ilg2016flownet, sun2018pwcnet, teed20raft:} are highly effective.
In particular, RAFT~\cite{teed20raft:} showed the effectiveness of an all-pairs inter-frame correlation volume as an encoding, which is essentially an attention map.
All-pairs intra-frame correlations were subsequently shown to help resolve correspondence ambiguities \cite{jiang2021learning}.
% Optical flow has been used in many networks as an explicit motion representation, invariant to appearance and texture and therefore allowing the network to focus on low-level motion statistics~\cite{Sevilla_Lara_2019}.
% Other works have proposed cheaper or better alternatives to optical flow as a motion representation~\cite{piergiovanni2018representation, fan2018end, gao2018im2flow, Zhang_2016, Zhu_2019, Sevilla_Lara_2019, shou2019dmcnet, Sun_2018}. 
% Optical flow has also been used a complementary view for unsupervised image~\cite{mahendran2018cross} and video~\cite{Han2020SelfsupervisedCF, piergiovanni2020evolving} representation learning, or improved supervised action recognition by using flow prediction as an auxiliary task~\cite{Ng_2018}. 
For longer-range correspondences, object tracking by repeated detection~\cite{ramanan05strike} and data association can be used.
In contrast to these approaches, our work does not explicitly establish temporal correspondences, but facilitates implicit correspondence learning via trajectory attention.
Jabri \etal~\cite{jabri_space_time} estimate correspondences in a similar way, framing the problem as a contrastive random walk on a graph and apply explicit guidance via a cycle consistency loss.
%where transition probabilities correspond to correspondence probabilities.
Incorporating such guidance into a video transformer is an interesting direction.

\section{Trajectory Attention for Video Data}\label{s:method}

Our goal is to modify the attention mechanism in transformers to better capture the information contained in videos.
Consider an input video $I \in \mathbb{R}^{T' \times 3 \times H\times W}$ consisting of $T'$ frames of resolution $H\times W$.
As in existing video transformer models~\cite{bertasius2021spacetime, arnab2021vivit}, we pre-process the video into a sequence of $ST$ tokens $\bx_{st} \in \reals^D$, for a spatial resolution of $S$ and a temporal resolution of $T$.
We use a cuboid embedding~\cite{arnab2021vivit, fan2021multiscale}, where disjoint spatio-temporal cubes from the input volume are linearly projected to $\reals^D$ (equivalent to a 3D convolution with downsampling).
We also test an embedding of disjoint image patches~\cite{dosovitskiy2021an}.
A learnable positional encoding $\be \in \reals^D$ is added to the video embeddings for spatial and temporal dimensions separately, resulting in the code $\bz_{st} = \bx_{st} + \be_s^s + \be_t^t$.
Finally, a learnable classification token $\bz_\text{cls}$ is added to the sequence of tokens, like in the BERT Transformer~\cite{bert18_naccl}, to reason about the video as a whole.
For clarity, we elide the classification token from our treatment in the sequel.

We now have a set of tokens that form the input to a sequence of transformer layers that, as in ViT~\cite{dosovitskiy2021an}, consist of Layer Norm (LN) operations~\cite{ba2016layer}, multi-head attention (MHA)~\cite{vaswani17attention}, residual connections~\cite{KaimingHe16}, and a feed-forward network (MLP):
\begin{equation}\label{e:transformer-recursion}
    \by = \text{MHA}(\text{LN}(\bz)) + \bz; \hspace{12pt} \bz' = \text{MLP}(\text{LN}(\by)) + \by.
\end{equation}
In the next section, we shall focus on a single head of the attention operation, and demonstrate how self-attention can realize a suitable inductive bias for video data.
For clarity of exposition, we abuse the notation slightly, neglecting the layer norm operation and using the same dimensions for single-head attention as for multi-head attention.

\subsection{Video self-attention}\label{s:attention}

\begin{figure}
  \centering
  \includegraphics[width=11cm,height=7cm,keepaspectratio]{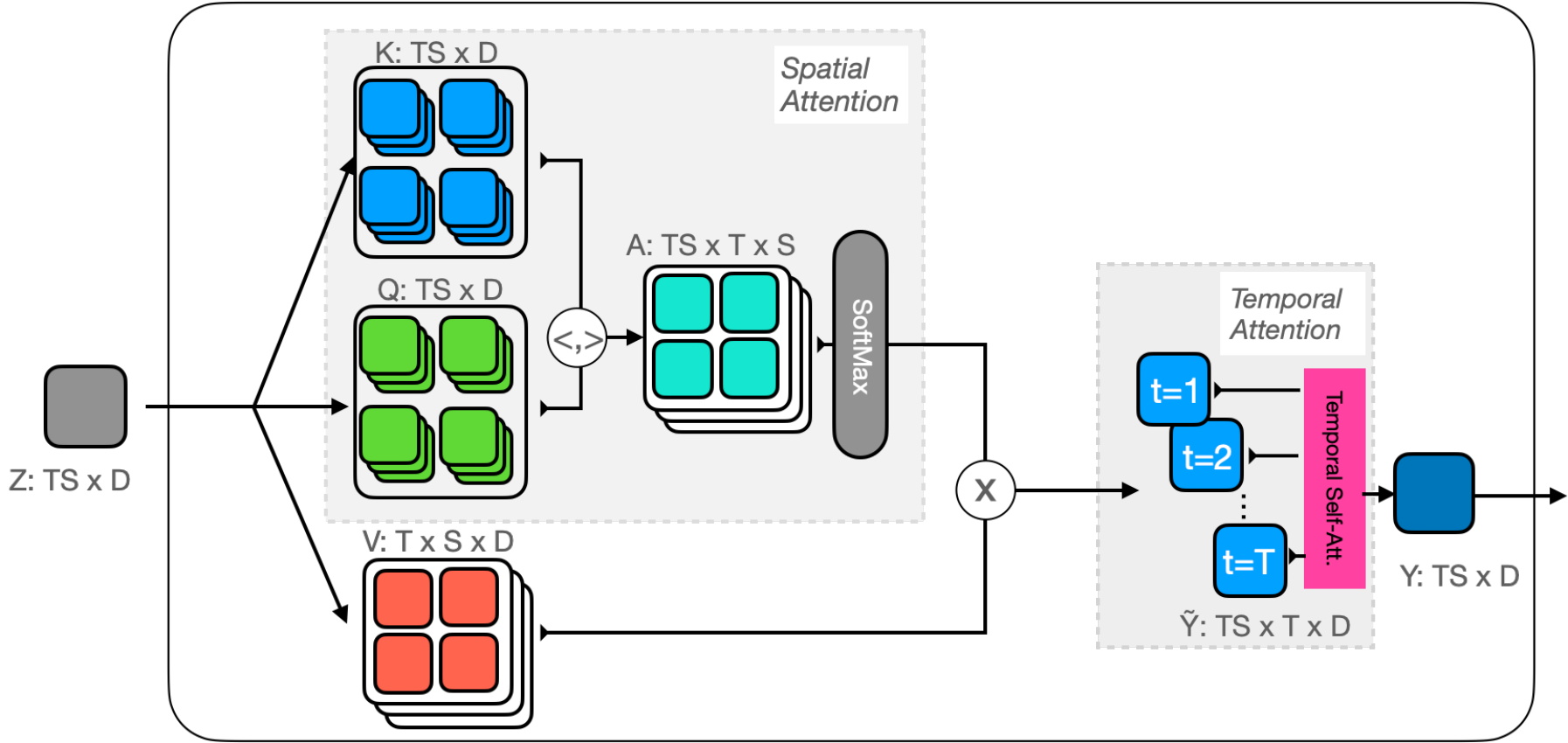}
  \caption{
      \textbf{Trajectory attention flowchart.}
      We divide the attention operation into two stages: the first forming a set of $ST$ trajectory tokens for every space-time location $st$---a spatial attention operation between pairs of frames---and the second pooling along these trajectories with a 1D temporal attention operation.
      In this way, we accumulate information along the motion paths of objects in the video.
    %   The first part computes intra-frame and inter-frame spatial attention for all queries, and the second part computes temporal attention along the motion-warped features for all queries.
      The softmax operations are computed over the last dimension.
  }\label{fig:attention}\vspace{-1em}
\end{figure}

The self-attention operation begins by forming a set of query-key-value vectors $\bq_{st},\bk_{st},\bv_{st} \in \reals^D$, one for each space-time location $st$ in the video.
These are computed as linear projections of the input $\bz_{st}$, that is, $\bq_{st} = \bW_q \bz_{st}$, $\bk_{st} = \bW_k \bz_{st}$, and $\bv_{st} = \bW_v \bz_{st}$, for projection matrices $\bW_i \in \reals^{D \times D}$.
A direct application of attention across space-time (called \emph{joint space-time attention}~\cite{bertasius2021spacetime,arnab2021vivit}) computes:
\begin{equation}\label{e:st-pooling}
\mathbf{y}_{st}
=
\sum_{s't'}
\mathbf{v}_{s't'} \cdot
\frac
{\exp\langle \mathbf{q}_{st}, \mathbf{k}_{s't'}\rangle}
{\sum_{\bar s\bar t}\exp \langle \mathbf{q}_{st}, \mathbf{k}_{\bar s\bar t}\rangle}.
\end{equation}
In this way, each query $\mathbf{q}_{st}$ is compared to all keys $\mathbf{k}_{s't'}$ using dot products, the results are normalized using the softmax operator, and the weights thus obtained are used to average the values corresponding to the keys.
Compared to a standard transformer, we have omitted for brevity the softmax temperature parameter $D^{1/2}$ and instead assume that the queries and keys have been divided by $D^{1/4}$.

One issue with this formulation is that it has quadratic complexity in both space and time, \ie, $\cO(S^2T^2)$.
An alternative is to restrict attention to either space or time (called \emph{divided space-time attention}):
\begin{equation}\label{s:s-or-t-pooling}
    \mathbf{y}_{st}
    =
    \sum_{s'}
    \mathbf{v}_{s't} \cdot
    \frac
    {\exp\langle \mathbf{q}_{st}, \mathbf{k}_{s't}\rangle}
    {\sum_{\bar s}\exp \langle \mathbf{q}_{st}, \mathbf{k}_{\bar st}\rangle}
    ~~\text{(space)};
    ~~~~
    \mathbf{y}_{st}
    =
    \sum_{t'}
    \mathbf{v}_{st'} \cdot
    \frac
    {\exp\langle \mathbf{q}_{st}, \mathbf{k}_{st'}\rangle}
    {\sum_{\bar t}\exp \langle \mathbf{q}_{st}, \mathbf{k}_{s\bar t}\rangle}
    ~~\text{(time)}.
\end{equation}
This reduces the complexity to $\cO(S^2T)$ and $\cO(ST^2)$, respectively, but only allows the model to analyse time and space independently.
This is usually addressed by interleaving~\cite{bertasius2021spacetime} or stacking~\cite{arnab2021vivit} the two attention modules in a sequence.

% Different to both of these approaches, we perform attention \emph{along trajectories}.
Different to both of these approaches, we perform attention \emph{along trajectories}, the probabilistic path of a token between frames.\footnote{Here, we refer to the trajectory as the motion between pairs of frames, rather than a multi-frame path.}
% For each space-time location $st$ (the trajectory `reference point') and corresponding query $\bq_{st}$, we construct a set of trajectory tokens $\tilde{\mathbf{y}}_{stt'}$.
For each space-time location $st$ (the trajectory `reference point') and corresponding query $\bq_{st}$, we construct a set of trajectory tokens $\tilde{\mathbf{y}}_{stt'}$, representing the pooled information weighted by the trajectory probability.
The trajectory extends for the duration of the video sequence and its tokens $\tilde{\mathbf{y}}_{stt'} \in \mathbb{R}^D$ at different times $t'$ are given by:
\begin{equation}
\tilde{\mathbf{y}}_{stt'}
=
\sum_{s'}
\mathbf{v}_{s't'} \cdot
\frac
{\exp\langle \mathbf{q}_{st}, \mathbf{k}_{s't'}\rangle}
{\sum_{\bar s}\exp \langle \mathbf{q}_{st}, \mathbf{k}_{\bar st'}\rangle}.
\end{equation}
Note that the attention in this formula is applied spatially (index $s$) and independently for each frame.
Intuitively, this pooling operation implicitly seeks the location of the trajectory at time $t'$ by comparing the trajectory query $\mathbf{q}_{st}$ to the keys $\mathbf{k}_{s't'}$ at time $t'$.

% Once the trajectories are computed, we further pool them across time. 
Once the trajectories are computed, we further pool them across time to reason about intra-frame information/connections.
To do so, the trajectory tokens are projected to a new set of queries, keys and values as usual:
\begin{equation}
    \tilde{\mathbf{q}}_{st} = \tilde{\mathbf{W}}_q\,
    \tilde{\mathbf{y}}_{stt},
    ~~~
    \tilde{\mathbf{k}}_{stt'} = \tilde{\mathbf{W}}_k\,
    \tilde{\mathbf{y}}_{stt'},
    ~~~
    \tilde{\mathbf{v}}_{stt'} = \tilde{\mathbf{W}}_v\,
    \tilde{\mathbf{y}}_{stt'}.
\end{equation}
Like $\mathbf{q}_{st}$ before, the updated reference query $\tilde{\mathbf{q}}_{st}$ corresponds to the trajectory reference point $st$ and contains information spatially-pooled from across the reference frame $t$.
This new query is used to pool across the new time (trajectory) dimension by applying 1D attention:
\begin{equation}
    \mathbf{y}_{st}
    =
    \sum_{t'}
    \tilde{\mathbf{v}}_{stt'} \cdot
    \frac
    {\exp\langle \tilde{\mathbf{q}}_{st}, \tilde{\mathbf{k}}_{stt'}\rangle}
    {\sum_{\bar t}\exp \langle \tilde{\mathbf{q}}_{st}, \tilde{\mathbf{k}}_{st\bar t}\rangle}.
\end{equation}
Like joint space-time attention, our approach has quadratic complexity in both space and time, $\cO(S^2 T^2)$, so has no computational advantage and is in fact slower than divided space-time attention.
However, we demonstrate better accuracy than both joint and divided space-time attention mechanisms.
We also provide fast approximations in \secref{s:approx}.
A flowchart of the full trajectory attention operation is shown in tensor form in \figref{fig:attention}.

\subsection{Approximating attention}\label{s:approx}

To complement our trajectory attention, we also propose an approximation scheme to speed up calculations.
This scheme is generic and applies to any attention-like pooling mechanism.
We thus switch to a generic transformer-like notation to describe it.
Namely, consider query-key-value matrices
$
\mathbf{Q},
\mathbf{K},
\mathbf{V}
\in \reals^{D\times N}
$
such that the query-key-value vectors are stored as columns
$
\mathbf{q}_i,
\mathbf{k}_i,
\mathbf{v}_i
\in \reals^{D}
$
in these matrices.

In order to obtain an efficient decomposition of the attention operator, we will rewrite it using a probabilistic formulation.
Let $A_{ij}\in \{0,1\}$ be a categorical random variable indicating whether the $j$th input (with key vector $\bk_{j} \in \reals^{D}$) is assigned to the $i$th output (with query vector $\bq_{i} \in \reals^{D}$), with $\sum_{j}A_{ij}=\ones$.
The attention operator uses a parametric model of the probability of this event based on the multinomial logistic function, \ie, the softmax operator $\softmax(\cdot)$:%
\footnote{
I.e.
$
[\softmax(\mathbf{z})]_i
=
\exp (z_i/\sqrt{D}) /
\sum_j \exp (z_j / \sqrt{D}).
$
For matrix inputs, the sum is over the columns.
}
\begin{equation}
P(A_{i:}) = \softmax(\bq_{i}\transpose \bK),\label{eq:prob-attn}
\end{equation}
where the subscript ${:}$ denotes a full slice of the input tensor in that dimension.
We now introduce the latent variables $U_{\ell j}\in \{ 0,1\}$, which similarly indicate whether the $j$th input is assigned to the $\ell$th \emph{prototype}, an auxiliary vector which we denote by $\bp_{\ell} \in \reals^{D}$.
We can use the laws of total and conditional probability to obtain:
\begin{equation}
P(A_{ij})=\sum\nolimits_{\ell}P(A_{ij} \mid U_{\ell j})P(U_{\ell j}).\label{eq:prob-attn-decomposition}
\end{equation}
Note that the latent variables that we chose are independent of the inputs (keys).
They use the same parametric model, but with parameters $\bP \in \reals^{D \times R}$ (the concatenated prototype vectors~$\bp_\ell$):\linebreak
$
P(U)=\softmax(\bP\transpose \bK).
$
\Eqnref{eq:prob-attn-decomposition} is \emph{exact}, even under the parametric model for $P(U)$, though the corresponding true distribution $P(A \mid U)$ is intractable. We now \emph{approximate} the conditional probability $P(A \mid U)$ with a similar parametric model:
\begin{equation}
\tilde{P}(A \mid U)=\softmax(\bQ\transpose \bP),\label{eq:prob-attn-conditional}
\end{equation}
where $\bQ \in \reals^{D \times N}$ concatenates all query vectors horizontally.
Substituting equations~\ref{eq:prob-attn}--\ref{eq:prob-attn-conditional} we write the full approximate attention $\tilde{\!\attn}$, multiplied by an arbitrary matrix $\bV$ (which in the case of a transformer contains the values of the key--value pairs stacked as rows):
\begin{equation}
\tilde{P}(A)\bV=\softmax(\bQ\transpose \bP)\left(\softmax(\bP\transpose \bK)\bV\right).\label{eq:prob-attn-approx}
\end{equation}

% Notes: $q \in \reals^{d \times m}$, $k \in \reals^{d \times n}$, $v \in \reals^{d \times n}$, $p \in \reals^{d \times r}$, $A \in \reals^{m \times n}$, $U \in \reals^{r \times n}$.

\paragraph{Computational efficiency.}

One important feature of the approximation in \eqnref{eq:prob-attn-approx} is that it can be computed in two steps.
First the values $\bV$ are multiplied by a prototypes-keys attention matrix $\softmax(\bP\transpose \bK)\in\reals^{R\times N}$, which can be much smaller than the full attention matrix $\softmax(\bQ\transpose \bK)\in\reals^{N\times N}$ (\eqnref{eq:prob-attn}), \ie, $R \ll N$.
Finally, this product is multiplied by a queries-prototypes attention matrix $\softmax(\bQ\transpose \bP)\in\reals^{N\times R}$, which is also small.
This allows us to sidestep the quadratic dependency of full attention over the input and output size ($\cO(N^2)$), replacing it with linear complexity ($\cO(N)$) as long as $R$ is kept constant.

\paragraph{Prototype selection.}
% The attention matrix is approximated using intermediate landmarks taken from the queries and keys.

% Greedy as-orthogonal-as-possible prototype selection:
The aim for prototype-based attention approximation schemes is to use as few prototypes as possible while reconstructing the attention operation as accurately as possible. As such, it behooves us to select prototypes efficiently. We have two priorities for the prototypes: to dynamically adjust to the query and key vectors so that their region of space is well-reconstructed, and to minimize redundancy. The latter is important because the relative probability of a query--key pair may be over-estimated if many prototypes are clustered near that query and key.
To address these criteria, we incrementally build a set of prototypes from the set of queries and keys such that a new prototype is maximally orthogonal to the prototypes already selected, starting with a query or key at random.
This greedy strategy is dynamic, since it selects prototypes from the current set of queries and keys, and has high entropy, since it preferences well-separated prototypes. % with respect to direction.
Moreover, it balances speed and performance by using a greedy strategy, rather than finding a globally-optimal solution to the maximum entropy sampling problem~\cite{shewry1987maximum}, making it suitable for use in a transformer.
% Shewry, M. C. and Wynn, H. P. (1987). Maximum entropy sampling
% Tease out the connection to the maximum entropy sampling problem? What we do is similar to maximising the determinant of the correlation matrix (although we use L_inf rather than L_2^2 wrt the dot product)

%Temporally shared prototypes:
Na{\"\i}vely applying prototype-based attention approximation techniques to video transformers would involve creating a unique set of prototypes for each frame in the video.
However, additional memory savings can be realized by sharing prototypes across time.
Since there is significant information redundancy between frames, video data is opportune for compression via temporally-shared prototypes. 
% This is likely to be a reasonable choice, since many scene elements in a frame persist across time, depending on the frame rate, camera motion, and scene motion. This information redundancy is opportune for compression via temporally-shared prototypes.

\paragraph{Orthoformer algorithm.}

The proposed approximation algorithm is outlined in \algoref{alg:proposed_approx}.
The attention matrix is approximated using intermediate prototypes, selected as the most orthogonal subset of the queries and keys, given a desired number of prototypes $R$.
% This selection procedure involves taking the dot product of $R(M+N)$ $D$-vectors.
To avoid a linear dependence on the sequence length $N$, we first randomly subsample $cR$ queries and keys, for a constant $c$, before selecting the most orthogonal subset, resulting in a complexity quadratic in the number of prototypes $\cO(R^2)$.
The algorithm then computes two attention matrices, much smaller than the original problem, and multiplies them with the values.
%
% Comparison with Nystromformer:
The most related approach in the literature is Nystr\"omformer~\cite{xiong2021nystromformer} attention, outlined in Algorithm~\ref{alg:nystromformer}.
This approach involves a pseudoinverse to attenuate the effect of near-parallel prototypes, has more operations, and a greater memory footprint.
% In particular, an iterative approximation to the pseudoinverse is used, requiring $4N_\text{iter}$ $(R \times R)$-matrix multiplications for $N_\text{iter}$ iterations.
% Taking the exact pseudoinverse, or increasing the number of iterations, degrades the performance.
% fewer landmarks, fewer operations, reduced memory requirements (unfortunately more sequential, so slower)

% Complexity analysis:
% IterativeInverse: $4n_\text{iter} \, (r \times r)$ matrix multiplications, $3n_\text{iter}$ matrix additions
% OrthoSelection: \Comment{$r \, ((m+n) \times d) \times (d \times 1)$ matrix multiplications}

\begin{minipage}[t]{.48\textwidth}
\vspace{-6mm}
\begin{algorithm}[H]\footnotesize
\begin{algorithmic}[1]
% \Require $\bQ$, $\bK$, $\bV$, $R$
% \Ensure $\bY$
\State $\bP \gets \text{MostOrthogonalSubset}(\bQ, \bK, R)$
\State $\bOmega_1 = \softmax(\bQ\transpose \bP / \sqrt{D})$
\State $\bOmega_2 = \softmax(\bP\transpose \bK / \sqrt{D})$
\State $\bY = \bOmega_1 (\bOmega_2 \bV)$
\end{algorithmic}
\caption{Orthoformer (proposed) attention}\label{alg:proposed_approx}
\end{algorithm}
\end{minipage}\hfill
\begin{minipage}[t]{.48\textwidth}
\vspace{-6mm}
\begin{algorithm}[H]\footnotesize
\begin{algorithmic}[1]
% \Require $\bQ$, $\bK$, $\bV$, $R$, $N_\text{iter}$
% \Ensure $\bY$
\State $\bP_q, \bP_k \gets \text{SegmentMeans}(\bQ, \bK, R)$
\State $\bOmega_1 = \softmax(\bQ\transpose \bP_k / \sqrt{D})$
\State $\bOmega_2^{-1} = \text{IterativeInverse}(\softmax(\bP_q\transpose \bP_k / \sqrt{D}), N_\text{iter})$
\State $\bOmega_3 = \softmax(\bP_q\transpose \bK / \sqrt{D})$
% \State $\bOmega_2^{-1} = \text{IterativeInverse}(\bOmega_2, N_\text{iter})$
\State $\bY = \bOmega_1 \left(\bOmega_2^{-1} \left(\bOmega_3 \bV \right)\right)$
\end{algorithmic}
\caption{Nystr\"omformer~\cite{xiong2021nystromformer} attention}\label{alg:nystromformer}
\end{algorithm}
\vspace{-7mm}
\end{minipage}

\subsection{The Motionformer model}\label{s:model}

\begin{table}[t]\footnotesize
	\centering
	\caption{\textbf{Comparison of recent video transformer models.} We show the different design choices of recent video transformer models and how they compare to our proposed Motionformer model.
% 	Trajectory attention is the key difference between these models.
	}
	\begin{tabularx}{\textwidth}{l c c C C}
		\toprule
		Model & Base Model & Attention & Pos. Encoding & Tokenization\\ 
		\midrule
		TimeSformer~\cite{bertasius2021spacetime}  & ViT-B & Divided Space--Time & Separate & Square   \\
		ViViT~\cite{arnab2021vivit}  &  ViT-L & Joint/Divided Space--Time & Joint & Cubic \\
		\midrule
		\textbf{Motionformer} &  ViT-B & Trajectory & Separate & Cubic \\ 
 		%\textbf{Motionformer-H} &  ViT-L & Trajectory (Orthoformer) & Separate & Cubic \\ 
		\bottomrule
	\end{tabularx}
	\vspace{-3mm}
	\label{tab:model_compare}
\end{table}

Our full video transformer model builds on previous work, as shown in \tabref{tab:model_compare}.
In particular, we use the ViT image transformer model~\cite{dosovitskiy2021an} as the base architecture, the separate space and time positional encodings of TimeSformer~\cite{bertasius2021spacetime}, and the cubic image tokenization strategy as in ViViT~\cite{arnab2021vivit}.
These design choices are ablated in \secref{s:experiments}.
The crucial difference for our model is the trajectory attention mechanism, with which we demonstrate greater empirical performance than the other models.
% We also show that the memory and time requirements can be significantly reduced without a large reduction in performance by approximating trajectory attention using the Orthoformer algorithm.

\section{Experiments}\label{s:experiments}

% \subsection{Datasets and implementation details}\label{s:experiments:datasets}

\paragraph{Datasets. \label{s:experiments:datasets}}
\textbf{Kinetics}~\cite{kinetics} is a large-scale video classification dataset consisting of short clips collected from YouTube, licensed by Google under Creative Commons.
As it is a dataset of human actions, it potentially contains personally identifiable information such as faces, names and license plates.
% We train and evaluate on Kinetics-400 and the larger Kinetics-600.
%
\textbf{Something--Something V2}~\cite{Goyal_2017} is a video dataset containing more than 200,000 videos across 174 classes, with a greater emphasis on short temporal clips.
In contrast to Kinetics, the background and objects remain consistent across different classes, and therefore models have to reason about fine-grained motion signals.
We verified the importance of temporal reasoning on this dataset by showing that a single frame model gets significantly worse results, a decrease of $39\%$ top-1 accuracy.
In contrast, a drop of only $7\%$ is seen on the Kinetics-400 dataset, showing that temporal information is much less relevant there.
We obtained a research license for this data from \url{https://20bn.com}; the data was collected with consent.
\textbf{Epic Kitchens-100}~\cite{damen2020rescaling} is an egocentric video dataset capturing daily kitchen activities.
% Each video is labelled with a ``verb'' (97 classes) and a ``noun'' (300 classes).
The highest scoring verb and action pair predicted by the network constitutes an action, for which we report top-1 accuracy.
The data is licensed under Creative Commons and was collected with consent by the Epic Kitchens teams.

\paragraph{Implementation details. \label{s:experiments:imp_details}}

We follow a standard training and augmentation pipeline~\citep{arnab2021vivit}, as detailed in the appendix.
%
% \paragraph{Preprocessing.}\label{s:experiments:imp_details}
% During training, we randomly sample clips of size $16$\x$224$\x$224$ at a rate of $1/4$ from $30$ FPS videos, thereby giving an effective temporal resolution of just over $2$ seconds.
% We normalize the inputs with mean and standard deviation 0.5, rescaling in the range $[-1, 1]$.
% We use standard video augmentations such as random scale jittering, random horizontal flips and color jittering.
% For smaller datasets such as Something--Something V2 and Epic-Kitchens, we additionally apply rand-augment~\cite{Cubuk2019RandAugmentPD} and mixup~\cite{zhang2018mixup}.
% During testing, we uniformly sample 10 clips per video and apply a 3 crop evaluation~\cite{slowfast}.
%
% \paragraph{Architecture.}
For ablations, our default Motionformer model is the Vision Transformer Base architecture~\cite{dosovitskiy2021an} (ViT/B), pretrained on ImageNet-21K~\cite{deng2009imagenet}, patch-size $2$\x$16$\x$16$ with central frame initialization~\cite{arnab2021vivit}, separate space-time positional embedding and our trajectory attention.
The base architecture has 12 layers, 12 attention heads, and an embedding dimension of 768.
Our default Motionformer model operates on $16$\x$224$\x$224$ videos with temporal stride 4 i.e. temporal extent of 2s. 
For comparisons with state-of-the-art, we report results on two additional variants: Motionformer-HR, which has a high spatial resolution ($16$\x$336$\x$336$ videos with temporal stride 4 i.e. temporal extent of 2s), and Motionformer-L, which has a long temporal range ($32$\x$224$\x$224$ videos with temporal stride 3 i.e. temporal extent of 3s).
Experiments with the large ViT architecture are deferred to the appendix.
%and Motionformer-H, which uses the large ViT architecture with a high spatial resolution and operates on $16$\x$336$\x$336$ videos.

% \paragraph{Training.}
% For all datasets, we use the AdamW~\cite{loshchilov2018fixing} optimizer with weight decay $5\times10^{-2}$, a batch size per GPU of 4, label smoothing~\cite{Inceptionv3} with alpha $0.2$ and mixed precision training~\cite{micikevicius2018mixed}.
% For Kinetics-400/600 and Something-Something V2, we train for 35 epochs, with an initial learning rate of $10^{-4}$, which we decay by 10 at epochs 20, 30.
% As Epic-Kitchens is a smaller dataset, we use a longer schedule and train for $50$ epochs with decay at $30$ and $40$.

\subsection{Ablation studies}

% \subsubsection{Inputs}

\paragraph{Input: tokenization.}

We consider the effect of different input tokenization approaches for both joint and trajectory attention on Kinetics-400 (K-400) and Something--Something V2 (SSv2) in~\tabref{tab:ablation-input:abl_pos_encoding}.
For patch tokenization ($1$\x$16$\x$16$), we use inputs of size $8$\x$224$\x$224$, while for cubic~\cite{arnab2021vivit, fan2021multiscale} tokenization ($2$\x$16$\x$16$), we use inputs of size $16$\x$224$\x$224$ to ensure that the model has the same number of input tokens over the same temporal range of 2 seconds.
For both attention types, we see that cubic tokenization gives a $1\%$ accuracy improvement over square tokenization on SSv2, a dataset for which temporal information is critical.
Furthermore, our proposed trajectory attention using cubic tokenization outperforms joint space-time attention on both datasets.

\paragraph{Input: positional encoding.}

Here, we ablate using a joint or separate~\cite{fan2021multiscale} (default) space-time positional encoding in~\tabref{tab:ablation-input:abl_pos_encoding}.
Similar to the results for input tokenization, the choice of positional encoding is particularly important for the fine-grained motion dataset, SSv2.
Since joint space-time attention treats tokens in the space-time volume equally, it benefits particularly from separating the positional encodings, allowing it to differentiate between space and time dimensions, with a $4\%$ improvement on SSv2 over joint space-time encoding.
Our proposed trajectory attention elicits a more modest improvement of $1\%$ from using separated positional encodings on SSv2, and outperforms joint space-time attention in both settings on both datasets.

\begin{table}[t]\footnotesize%\scriptsize
    \caption{\textbf{Input encoding ablations:} Comparison of input tokenization and positional encoding design choices. We report GFLOPS and top-1 accuracy ($\%$) on K-400 and SSv2.}\label{tab:ablation}
    \setlength{\tabcolsep}{2pt}
	\begin{subtable}[t]{.49\linewidth}\centering
    	\caption{Cubic tokenization works best for trajectory attn.}
    	\label{tab:ablation-input:abl_tokenization}
    	\begin{tabularx}{\linewidth}{l l C C C}
	\toprule
	Attention & Tokenization        & GFlops & K-400 & SSv2  \\ 
	\midrule
	Joint ST &  Square (1$\times$16$^2$)  &  179.7  & 79.4 & 63.0    \\[-0.5pt]
	      &  Cubic (2$\times$16$^2$)   &  180.6  & 79.2 & 64.0  \\[-0.7pt]
	\midrule
	\bf Trajectory & Square (1$\times$16$^2$)  &  368.5  & 79.4 & 65.8   \\[-0.5pt]
	          & Cubic (2$\times$16$^2$)  &   369.5 & \bf 79.7 & \bf 66.5 \\[-0.7pt]
	 \bottomrule
\end{tabularx}
	\end{subtable}
	\hfill
	\begin{subtable}[t]{.49\linewidth}\centering
    	\caption{Trajectory attn. works well with both encodings.}
    	\label{tab:ablation-input:abl_pos_encoding}
    	\begin{tabularx}{\linewidth}{l l C C C}
	\toprule
	Attention & Pos. Encoding          & GFlops & K-400 & SSv2  \\ 
	\midrule
	Joint ST &  Joint ST  &  180.6  & 79.1 & 60.8    \\ 
	      &  Separate ST~\cite{fan2021multiscale}  &  180.6  & 79.2 & 64.0  \\ 
	\midrule
	\bf Trajectory & Joint ST  &  369.5  & 79.6 & 65.8   \\ 
	          & Separate ST~\cite{fan2021multiscale}  &  369.5  & \bf 79.7 & \bf 66.5 \\ 
	 \bottomrule
\end{tabularx}
	\end{subtable}
% 	\begin{subtable}[t]{.5\linewidth}\centering
% 	    {\input{tabs/abl_approx_xformer}}	
% 	\end{subtable}
% 	\begin{subtable}[t]{.5\linewidth}\centering
% 	    {\input{tabs/abl_approx_landmarks_num}}	
% 	\end{subtable}
\vspace{-1.5em}
\end{table}

\begin{table}[t]\footnotesize%\scriptsize
\caption{\textbf{Orthoformer ablations:} We ablate various aspects of our Orthoformer approximation. E denotes exact attention and A denotes approximate attention. We report max CUDA memory consumption (GB) and top-1 accuracy ($\%$) on K-400 and SSv2.}
\label{tab:ablation-approx}
\setlength{\tabcolsep}{2pt}
	\begin{subtable}[t]{.49\linewidth}\centering
    	\caption{Orthoformer is competitive with Nystr\"om.}
    	\label{tab:ablation-approx:xformer}
    	\begin{tabularx}{\linewidth}{l l C c C}
	\toprule
	Attention & Approx. & Mem. & K-400 & SSv2  \\
	\midrule
	Trajectory (E)    &  N/A  &  7.4  & \bf 79.7 & \bf 66.5    \\ 
	\midrule
	Trajectory (A)
	                  & Performer     &  5.1  & 72.9 & 52.7    \\ 
	                  & Nystr\"omformer  &  3.8  & \textbf{77.5} & \textbf{64.0}  \\ 
	                  & \bf Orthoformer   &  3.6  & \textbf{77.5} & 63.8   \\
	 \bottomrule
\end{tabularx}
	\end{subtable}
	\hfill
	\begin{subtable}[t]{.49\linewidth}\centering
    	\caption{Selecting orthogonal prototypes is the best strategy.}
    	\label{tab:ablation-approx:landmark_sel}
    	\begin{tabularx}{\linewidth}{l l C C C}
%\label{tab:ablation-approx:land_sel}
	\toprule
	Attention & Selection & Mem. & K-400 & SSv2  \\
	\midrule
	Trajectory (E) &  N/A  &  7.4  & \bf 79.7 & \bf 66.5    \\ 
	\midrule
	Trajectory (A)  &  Seg-Means   &  3.6  & 75.8 & 60.3  \\  
	                   &  Random  &  3.6  & 76.5 & 62.5    \\ 
	                  &  \textbf{Orthogonal}  &  3.6  & \textbf{77.5} & \textbf{63.8}   \\ 
	 \bottomrule
\end{tabularx}
	\end{subtable} 
    \vfill
    \begin{subtable}[t]{.49\linewidth}\centering
        \caption{Approximation improves with more prototypes.}
    	\label{tab:ablation-approx:landmarks_num}
		\begin{tabularx}{\linewidth}{l c C C C}
	\toprule
	Attention & \# Prototypes & Mem. & K-400 & SSv2  \\
	\midrule
	Trajectory (E) &  N/A  &  7.4  & \bf 79.7 & \bf 66.5    \\ 
	\midrule
	Trajectory (A) &  16  &  3.1  & 73.9 & 59.2    \\ 
	                   &  64   &  3.3  & 74.9 & 63.0  \\ 
	                   & 128  &  3.6  & \textbf{77.5} & \textbf{63.8}   \\ 
	 \bottomrule
\end{tabularx}
	\end{subtable}
	\hfill
	\begin{subtable}[t]{.49\linewidth}\centering
    	\caption{Temporal sharing is the best strategy.}
    	\label{tab:ablation-approx:landmarks_shared}
		\begin{tabularx}{\linewidth}{l c C C C}
	\toprule
	Attention & Sharing & Mem. & K-400 & SSv2  \\
	\midrule
	Trajectory (E) &  N/A  &  7.4  & \bf 79.7 & \bf 66.5    \\ 
	\midrule
	Trajectory (A) &  \xmark  &  16.5  & 77.3 & 61.5    \\ 
	                   &  \checkmark     &  3.6  & \textbf{77.5} & \textbf{63.8}   \\
	 \bottomrule
\end{tabularx}
	\end{subtable}
\vspace{-10pt}
\end{table}

\begin{table}[t]\footnotesize
\centering
\caption{\textbf{Attention ablations:} We compare trajectory attention with alternatives and ablate its design choices. We report GFLOPS and top-1 accuracy ($\%$) on K-400 and SSv2. Att$_T$: temporal attention, Avg$_T$: temporal averaging, Norm$_{ST}$: space-time normalization, Norm$_{S}$: spatial normalization.}
\label{tab:ablation-attention}
\setlength{\tabcolsep}{6pt}
\begin{tabularx}{\linewidth}{l | c c | c c | C | C C}
	\toprule
	Attention  & Att$_T$ & Avg$_T$ & Norm$_S$ & Norm$_{ST}$ & GFLOPS & K-400 & SSv2  \\
	\midrule
	Joint Space-Time  & --  & --  & --  & --  & 180.6 & 79.2 & 64.0 \\
	Divided Space-Time & --  & --  & --  & --  & 185.8 & 78.5 &  64.2 \\
	\midrule
	            & \xmark  & \checkmark & \checkmark & \xmark \  & 180.6  &  76.0 &  60.0 \\
	            &  \checkmark  & \xmark & \xmark  & \checkmark  &  369.5  &  77.2 &  60.9 \\
	 Trajectory & \checkmark  & \xmark  & \checkmark  & \xmark & 369.5   & \bf 79.7 & \bf 66.5 \\
	 \bottomrule
\end{tabularx}

% \begin{tabular}{l c c c}
% 	\toprule
% 	Trajectory  & Flops. & K-400 & SSv2  \\
% 	\midrule
% 	Averaging & 180.6  & 76.0 & 60.0    \\ 
% 	Attention  &  369.5  & \bf 79.7 & \bf 66.5    \\ 
% % 	\midrule
% % 	Query (O)  &  45G  & XX & XX    \\ 
% % 	Query (Agg)  &  45G  & XX & XX   \\ 
% 	 \bottomrule
% \end{tabular}
\vspace{-1em}
\end{table}

% \subsubsection{Attention block}

\paragraph{Attention block: comparisons.}

We compare our proposed trajectory attention to joint space-time attention~\cite{arnab2021vivit}, and divided space-time attention~\cite{bertasius2021spacetime} in \tabref{tab:ablation-attention}.
Our trajectory attention (bottom row) outperforms both alternatives on the K-400 and SSv2 datasets.
While we see only modest improvements on the appearance cue-reliant K-400 dataset, our trajectory attention significantly outperforms ($+2\%$) the other approaches on the motion cue-reliant SSv2 dataset.
This dataset requires fine-grained motion understanding, something explicitly singled out by previous video transformer works~\cite{arnab2021vivit, bertasius2021spacetime} as a challenge for their models.
In contrast, our trajectory attention excels on this dataset, indicating that its motion-based design is able to capture some of this information.

\paragraph{Attention block: trajectory attention design.}

We ablate two design choices for our trajectory attention: the per-frame softmax normalization and the 1D temporal attention.
Unlike joint space-time attention, which normalizes the attention map over all tokens in space and time, trajectory attention normalizes independently per frame, allowing us to implicitly track the trajectories of query patches in time.
In row 5 of \tabref{tab:ablation-attention}, we ablate the benefits of this design choice.
We observe a reduction of $2.5\%$ on K-400 and $5.6\%$ on SSv2 by normalizing over space and time (Norm$_{ST}$) compared with normalizing over space alone (Norm$_{S}$).
In row 4, we show the benefit of using 1D temporal attention (Att$_T$) to aggregate temporal features, compared to average pooling (Avg$_T$).
We observe reductions of $3.7\%$ on K-400 and  $6.5\%$ on SSv2 when using average pooling instead of temporal attention applied to the motion trajectories, although it saves computing the additional query/key/value projections.

\subsection{Orthoformer approximated attention}
\begin{table}[tb]
    \setlength{\tabcolsep}{4pt}
    \footnotesize
    \centering
    \caption{\textbf{Comparison to the state-of-the-art on Long Range Arena benchmark.} GFLOPS and CUDA maximum Memory (MB) are reported for the ListOps task. Note that our algorithm achieves the best overall results with far fewer prototypes (64) than the other methods.}
    \label{tab:lra}
    \begin{tabularx}{\linewidth}{l C C C C c | c | c c}
    \toprule
    {Model} & {ListOps} & {Text} & {Retrieval} & {Image} & {Pathfinder} & {Avg}$\uparrow$ & GFLOPS$\downarrow$ & Mem.$\downarrow$\\ 
    \midrule
    Exact~\cite{vaswani17attention} & \underline{36.69} & 63.09 & 78.22  & 31.47 & 66.35 & \underline{55.16} & 1.21 & 4579 \\
    \midrule
    Performer-256~\cite{choromanski2021rethinking} & \underline{36.69} & 63.22 & \bf 78.98 & 29.39 & \bf 66.55 & 54.97 & \underline{0.49} & 885 \\
    Nyströmformer-128~\cite{xiong2021nystromformer} & \bf 36.90 & \underline{64.17} & \underline{78.67} & \textbf{36.16} & 52.32 & 53.64 & 0.62 & \underline{745} \\
    \midrule
    \bf{Orthoformer-64} & 33.87 & \bf 64.42 & 78.36 & \underline{33.26} & \underline{66.41} & \bf{55.26} & \bf 0.24 & \bf 344 \\
    \bottomrule
    \end{tabularx}
\vspace{-6mm}
\end{table}

\begin{table}
\caption{\textbf{Comparison to the state-of-the-art on video action recognition.} We report GFLOPS and top-1 ($\%$) and top-5 ($\%$) video action recognition accuracy on K-400/600, and SSv2. On Epic-Kitchens, we report top-1 ($\%$) action (A), verb (V), and noun (N) accuracy.}\label{tab:sota}
	\begin{subtable}[t]{.49\linewidth}\scriptsize
	    \centering
	    \setlength{\tabcolsep}{3pt}
	    \caption{\bf{Something--Something V2}}
		\begin{tabularx}{\linewidth}{@{}l c C C r@{}}
\toprule
{Model} & Pretrain & {Top-1} & {Top-5} & GFLOPs \x views \\ 
\midrule
SlowFast~\cite{slowfast} & K-400 & 61.7 & - & 65.7\x3\x1\\
TSM~\cite{Lin_2019} & K-400 & 63.4 & 88.5 & 62.4\x3\x2 \\
STM~\cite{jiang2019stm} & IN-1K & 64.2 & 89.8 & 66.5\x3\x10 \\
MSNet~\cite{kwon2020motionsqueeze} & IN-1K & 64.7 & 89.4 & 67\x1\x1 \\
TEA~\cite{li2020tea} & IN-1K & 65.1 & - & 70\x3\x10 \\
bLVNet~\cite{fan2019blvnet} & IN-1K & 65.2 & 90.3 & 128.6\x3\x10 \\
\midrule
VidTr-L~\cite{li2021vidtr} & {\scriptsize IN-21K+K-400} &  60.2 & - & 351\x3\x10 \\ 
Tformer-L~\cite{bertasius2021spacetime} & IN-21K & 62.5 & - & 1703\x3\x1 \\
ViViT-L~\cite{arnab2021vivit} & {\scriptsize IN-21K+K-400}  & 65.4 & 89.8 & 3992\x4\x3 \\
MViT-B~\cite{fan2021multiscale} & K-400  &  67.1 &  90.8 & 170\x3\x1 \\
\midrule
\bf Mformer & {\scriptsize IN-21K+K-400} & 66.5 & 90.1 & 369.5\x3\x1 \\
\bf Mformer-L & {\scriptsize IN-21K+K-400} & \bf 68.1 & \bf 91.2 & 1185.1\x3\x1 \\
\bf Mformer-HR & {\scriptsize IN-21K+K-400}  & 67.1 & 90.6 & 958.8\x3\x1 \\
\bottomrule
\end{tabularx}

	\end{subtable}
	\hfill
    \begin{subtable}[t]{.49\linewidth}\scriptsize
        \centering
        \caption{\bf{Kinetics-400}}
        \label{tab:sota_kinetics400}
		\begin{tabularx}{\linewidth}{@{}l c C C r@{}}
\toprule
{Method} & Pretrain & {Top-1} & {Top-5} &  GFLOPs\x views \\ 
\midrule
%ARTNet~\cite{wang2017appearance} & - & 69.2 & 88.3 & 6.0 \\ 
I3D~\cite{I3D} & IN-1K & 72.1 & 89.3 & 108\x N/A \\
R(2+1)D~\cite{tran2018closer} & - & 72.0 & 90.0 & 152\x5\x23\\
% MFNet~\cite{lee2018motion} & IN-1K & 72.8 & 90.4 & N/A \\
% Inception-ResNet~\cite{bian2017revisiting} & ? & 73.0 & 90.9 & N/A\\
% bLVNet~\cite{fan2019blvnet} & IN-1K &  73.5 & 91.2 & 0.84\\
% $A^2$-Net~\cite{doubleattention} & ? & 74.6 & 91.5 & N/A \\
% TSM~\cite{Lin_2019} & IN-1K & 74.7 & N/A & 62.4 \x N/A\\
S3D-G~\cite{xie2018rethinking} & IN-1K & 74.7 & 93.4 & 142.8\x N/A \\
%Oct-I3D+NL~\cite{Chen_2019_ICCV} & IN-1K & 75.7 & N/A & 0.84\\
% D3D~\cite{Stroud_2020} & IN-1K & 75.9 & N/A & N/A \\
% GloRe~\cite{Chen_2019_CVPR} & IN-1K & 76.1 & N/A & N/A \\
% I3D+NL~\cite{XiaolongWang18} & IN-1K  & 77.7 & 93.3 & 10.8 \\
% ip-CSN~\cite{Tran_2019} & - & 77.8 & 92.8 & 109\x3\x10 \\ 
X3D-XL~\cite{Feichtenhofer_2020} & - & 79.1 & 93.9 & 48.4\x3\x10  \\
% CorrNet~\cite{Wang_2020} & - & 79.2 & N/A & 224\x3\x10 \\
% LGD-3D~\cite{Qiu_2019} & IN-1K & 79.4 & 94.4 & N/A\\ 
SlowFast~\cite{slowfast} & - & 79.8 & 93.9 & 234\x3\x10 \\
\midrule
VTN~\cite{neimark2021video} & IN-21K & 78.6 & 93.7 & 4218\x1\x1 \\ 
VidTr-L~\cite{li2021vidtr} & IN-21K & 79.1 & 93.9 & 392\x3\x10 \\ 
Tformer-L\cite{bertasius2021spacetime} & IN-21K  &  80.7 &  94.7 & 2380\x3\x1  \\
MViT-B~\cite{fan2021multiscale} & -  &  81.2 &  95.1 & 455\x3\x3  \\
ViViT-L~\cite{arnab2021vivit} & IN-21K  & \bf 81.3 & 94.7 & 3992\x3\x4 \\
\midrule
\bf Mformer & IN-21K & 79.7 & 94.2 & 369.5\x3\x10 \\
\bf Mformer-L & IN-21K & 80.2 & 94.8 & 1185.1\x3\x10 \\
%\bf Mformer-HR & IN-21K & \underline{80.7} & \underline{94.9} & 958.8\x3\x10 \\
\bf Mformer-HR & IN-21K & 81.1 & \bf{95.2} & 958.8\x3\x10 \\
%\bf Mformer-H & IN-21K & \textcolor{red}{XX} & \textcolor{red}{XX} & 2786.9\x3\x10 \\
\bottomrule
\end{tabularx}
	\end{subtable}
	\\
	\begin{subtable}[t]{.49\linewidth}\scriptsize
	    \centering
	    \caption{\bf{Epic-Kitchens}}
	    \begin{tabularx}{\linewidth}{@{}l c C C C@{}}
\toprule
{Method} & Pretrain & {A} & {V} & {N}  \\ 
\midrule
TSN~\cite{wang16temporal} & IN-1K & 33.2 & 60.2 & 46.0 \\ 
TRN~\cite{Zhou_2018} & IN-1K & 35.3 & 65.9 & 45.4 \\ 
TBN~\cite{kazakos2019epic} & IN-1K & 36.7 & 66.0 & 47.2 \\
TSM~\cite{Lin_2019} & IN-1K & 38.3 & \bf 67.9 & 49.0 \\
SlowFast~\cite{slowfast} & K-400  & 38.5 & 65.6 & 50.0 \\
\midrule
ViViT-L~\cite{arnab2021vivit} & {\scriptsize IN-21K+K-400}  & 44.0 & 66.4 & 56.8 \\
\midrule
\bf Mformer & {\scriptsize IN-21K+K-400} & 43.1 & 66.7 & 56.5 \\
\bf Mformer-L & {\scriptsize IN-21K+K-400} & 44.1 & 67.1 & 57.6 \\
\bf Mformer-HR & {\scriptsize IN-21K+K-400} & \bf 44.5 & \underline{67.0} & \bf 58.5 \\
\bottomrule
\end{tabularx}
	\end{subtable}
	\hfill
	\begin{subtable}[t]{.49\linewidth}\scriptsize
	    \centering
	    \setlength{\tabcolsep}{4pt}
	    \caption{\bf{Kinetics-600}}
		\begin{tabularx}{\linewidth}{@{}l c C C r@{}}
\toprule
{Model} & Pretrain & {Top-1} & {Top-5} & GFLOPs \x views \\ 
\midrule
AttnNAS~\cite{wang2020attentionnas} & - & 79.8 & 94.4 & - \\ 
LGD-3D~\cite{Qiu_2019} & IN-1K & 81.5 & 95.6 & - \\ 
SlowFast~\cite{slowfast} & - & 81.8 & 95.1 & 234\x3\x10 \\
X3D-XL~\cite{Feichtenhofer_2020} & - & 81.9 & 95.5 & 48.4\x3\x10 \\
\midrule
Tformer-HR~\cite{bertasius2021spacetime} & IN-21K  & 82.4 & 96.0 & 1703\x3\x1 \\
ViViT-L~\cite{arnab2021vivit} & IN-21K  &  83.0 & 95.7 & 3992\x3\x4 \\
MViT-B-24~\cite{fan2021multiscale} & -  &  \bf 83.8 & \bf 96.3 & 236\x1\x5 \\
\midrule
\bf Mformer & IN-21K & 81.6 & 95.6 & 369.5\x3\x10 \\
\bf Mformer-L & IN-21K & 82.2 & 96.0 & 1185.1\x3\x10 \\
\bf Mformer-HR & IN-21K & \underline{82.7} & 96.1 & 958.8\x3\x10 \\
\bottomrule
\end{tabularx}

	\end{subtable}
-\vspace{-1em}
\end{table}
\paragraph{Approximation comparisons.} In \tabref{tab:ablation-approx:xformer}, we compare our Orthoformer algorithm to alternative strategies: Nystr\"omformer~\cite{xiong2021nystromformer} and Performer~\cite{choromanski2021rethinking}.
Our algorithm performs comparably with Nystr\"omformer with a reduced memory footprint.
% Both Nystr\"omformer and our proposed Orthoformer significantly outperform Performer on video data, and achieve similar performance on both datasets.
%
In ~\tabref{tab:lra}, we also compare these attention mechanisms on the Long Range Arena benchmark~\cite{tay2021long} to show applicability to other tasks and data types.
Orthoformer is able to effectively approximate self-attention, outperforming the state-of-the-art despite using  far fewer prototypes (64) and so gaining significant computational and memory benefits.
% Similarly to video data, our Orthoformer approximation is able to effectively approximate self-attention, and outperforms both Nystr\"omformer and Performer, with significantly lower FLOPs and memory requirements.

\paragraph{Prototype selection.} A key part of our Orthoformer algorithm is the prototype selection procedure.
In \tabref{tab:ablation-approx:landmark_sel}, we ablate three prototype selection strategies: segment-means, random, and greedy most-orthogonal selection.
Segment-means, the strategy used in Nystr\"omformer, performs poorly because it can generate multiple parallel prototypes, which will over-estimate the relative probability of query--key pairs near those redundant prototypes.
In contrast, our proposed strategy of selecting the most orthogonal prototypes from the query and key set works the best across both datasets, because it explicitly minimises prototype redundancy with respect to direction.

\paragraph{Number of prototypes.} In ~\tabref{tab:ablation-approx:landmarks_num}, we show that Orthoformer improves monotonically as the number of prototypes is increased.
In particular, we see an average performance improvement of $4\%$ on both datasets as we increase the number of prototypes from 16 to 128.

\paragraph{Temporally-shared prototypes.} In ~\tabref{tab:ablation-approx:landmarks_shared}, we demonstrate the memory savings and performance benefits of sharing prototypes across time.
On SSv2, we observe a $2\%$ improvement in performance and a $5\times$ decrease in memory usage.
These gains may be attributed to the regularization effect of having prototypes leverage redundant information across frames.

\paragraph{Scaling transformer models with approximated trajectory attention.}\label{s:large}
The Orthoformer attention approximation algorithm allows us to train larger models and higher resolution inputs for a given GPU memory budget.
Here, we verify this, by training a large vision transformer model (ViT-L/16)~\cite{dosovitskiy2021an} with a higher resolution input ($336\times336$ pixels) on the Kinetics-400 dataset, using the Orthoformer approximation with 196 temporally-shared prototypes and the same schedule as the base model.
We use a fixed patch size (in pixels) for all models, and so the number of input tokens to the transformer scales with the square of the image resolution.
As shown in \tabref{tab:large_model}, this model achieves a competitive accuracy without fine-tuning the training schedule, hyperparameters or data augmentation strategy.
We expect that fine-tuning these on a validation set would greatly improve the model's performance, based on results from contemporary work~\cite{arnab2021vivit}.
Obviously such a parameter sweep is more time-consuming for these large models, however, these preliminary results are indicative that higher accuracies are attainable if these parameters were to be optimized.

% \begin{table}[t]%\footnotesize
% 	\centering
% 	\caption{\textbf{Can we train larger models using approximated trajectory attention?}
% 	We report top-1 and top-5 accuracy (\%) on the Kinetics-400 dataset of two variants of our Motionformer model: Motionformer-B and Motionformer-H.
% 	The former uses the base model with exact (E) trajectory attention, while the latter uses a much larger model (ViT-L) and a higher resolution input ($336\times336$ pixels) with approximate (A) trajectory attention, \ie, using Orthoformer.
% 	The larger model has better performance, despite no optimization of the training schedule, hyperparameters, and data augmentation schedule.
% 	The larger model also has far more parameters than the base model, and so unavoidably requires more GPU memory.
% 	Furthermore, for a fixed patch size (in pixels), the memory requirements for exact attention scale with the square of the input resolution.
% 	We reduce this to a linear relationship with the Orthoformer approximation, which allows us to fit the model on the GPU.
% 	}
% 	\begin{tabularx}{\textwidth}{l c c c c C C}
% 	\toprule
% 		Model & Base model & Params & Attention & Max memory (GB) & Top-1 & Top-5 \\ 
% 		\midrule
% 		Mformer-B &  ViT-B/224 & 109.1M & Trajectory (E) & \textbf{7.3} & 79.7 & 94.2 \\ 
%  		Mformer-H &  ViT-L/336 & 381.9M & Trajectory (A) & 22.2 & \textbf{80.0} & \textbf{94.5} \\ 
% 		\bottomrule
% 	\end{tabularx}
% 	\vspace{-3mm}
% 	\label{tab:large_model}
% \end{table}

\begin{table}[t]%\footnotesize
	\centering
	\caption{\textbf{Can we train larger models using approximated trajectory attention?}
	We report top-1 and top-5 accuracy (\%) on the Kinetics-400 dataset of two variants of our Motionformer model: Motionformer-B and Motionformer-H.
	The former uses the base model with exact (E) trajectory attention, while the latter uses a much larger model (ViT-L) and a higher resolution input ($336\times336$ pixels) with approximate (A) trajectory attention, \ie, using Orthoformer.
% 	The larger model has better performance, despite no optimization of the training schedule, hyperparameters, and data augmentation schedule.
% 	The larger model also has far more parameters than the base model, and so unavoidably requires more GPU memory.
% 	Furthermore, for a fixed patch size (in pixels), the memory requirements for exact attention scale with the square of the input resolution.
	We reduce this to a linear relationship with the Orthoformer approximation, which allows us to fit the model on the GPU.
	}
	\begin{tabularx}{\textwidth}{l c c c c C C}
	\toprule
		Model & Base model & Params & Attention & Max memory (GB) & Top-1 & Top-5 \\ 
		\midrule
		Mformer-B &  ViT-B/224 & 109.1M & Trajectory (E) & \textbf{7.3} & 79.7 & 94.2 \\ 
 		Mformer-H &  ViT-L/336 & 381.9M & Trajectory (A) & 22.2 & \textbf{80.0} & \textbf{94.5} \\ 
		\bottomrule
	\end{tabularx}
	\vspace{-3mm}
	\label{tab:large_model}
\end{table}

% \subsection{Qualitative analysis}

\subsection{Comparison to the state-of-the-art}
In \tabref{tab:sota}, we compare our method against the current state-of-the-art on four common benchmarking datasets: Kinetics-400, Kinetics-600, Something--Something v2 and Epic-Kitchens.
We find that our method performs favorably against current methods, even when compared against much larger models such as ViViT-L.
In particular, it achieves strong top-1 accuracy improvements of $1.0\%$ and $2.3\%$ for SSv2 and Epic-Kitchen Nouns, respectively.
These datasets require greater motion reasoning than Kinetics and so are a more challenging benchmark for video action recognition.
% differentiate between the models
% for which our model is particularly suited.
% and therefore our method is able to outperform both transformer-based and 3D-CNN based models by a significant margin.

\section{Conclusion \label{ref:conclusion}}

We have presented a new general-purpose attention block for video data that aggregates information along implicitly determined motion trajectories, lending a realistic inductive bias to the model.
We further address its quadratic dependence on the input size with a new attention approximation algorithm that significantly reduces the memory requirements, the largest bottleneck for transformer models.
With these contributions, we obtain state-of-the-art results on several benchmark datasets.
%
% Limitations:
Nonetheless, our approach inherits many of the limitations of transformer models, including poor data efficiency and slow training.
Specific to this work, trajectory attention has higher computational complexity than alternative attention operations used for video data. % higher memory and FLOP requirements
This is attenuated by the proposed approximation algorithm, with significantly reduced memory and computation requirements.
However, its runtime is bottlenecked by prototype selection, which is not easily parallelized.
%to the detriment of the runtime.
%, and so the runtime savings are not fully realized.

%Limitations: computational complexity (memory and time requirements); data efficiency still far from ideal; training is slow; greedy prototype selection for Orthoformer is not easily parallelizable
% data efficiency could be improved by making the motion learning more explicit, eg by using optical flow

\paragraph{Future work.}
There are many applications of trajectory attention beyond video action classification, such as those tasks where temporal context is highly important.
We see significant potential for using trajectory attention for tracking~\cite{henriques2014high}, temporal action localization~\cite{wu2019long, weinzaepfel2015learning} and online action detection~\cite{xu2021long, singh2017online}, among other settings, and leave these as avenues for future work.

\paragraph{Potential negative societal impacts.}

One negative impact of this research is the significant environmental impact associated with training transformers, which are large and compute-expensive models.
Compared to 3D-CNNs where the compute scales linearly with the sequence length, video transformers scale quadratically.
To mitigate this, we proposed an approximation algorithm with linear complexity that greatly reduces the computational requirements.
There is also potential for video action recognition models to be misused, such as for unauthorized surveillance. %, especially by autocratic regimes, which disproportionately affects minority and marginalized communities.
% The model is also imperfect, and any use of the technology should recognize that incorrect results are unlikely to be randomly distributed, but to reflect biases in the datasets.

\begin{ack}
We are grateful for support from the Rhodes Trust (M.P.), the European Research Council Starting Grant (IDIU 638009, D.C.), Qualcomm Innovation Fellowship (Y.A.), the Royal Academy of Engineering (RF201819/18/163, J.H.), and EPSRC Centre for Doctoral Training in Autonomous Intelligent Machines \& Systems (EP/L015897/1, M.P. and Y.A.). 
Funding for M.P. was received under his Oxford affiliation.
We thank Bernie Huang, Dong Guo, Rose Kanjirathinkal, Gedas Bertasius, Mike Zheng Shou, Mathilde Caron, Hugo Touvron, Benjamin Lefaudeux, Haoqi Fan, and Geoffrey Zweig from Facebook AI for their help, support, and discussion around this project.
We also thank Max Bain and Tengda Han from VGG for fruitful discussions.
\end{ack}

\clearpage

{\small\bibliographystyle{ieee_fullname}\bibliography{egbib,refs,vedaldi_general}}

\clearpage
\section{Appendix}

\subsection{Further experimental analysis and results}\label{s:analysis}

\subsubsection{Does trajectory attention make better use of motion cues?}\label{s:motion}

\begin{figure}[h]\centering
    \begin{subfigure}[]{0.5\textwidth}\centering
    	\includegraphics[width=\textwidth]{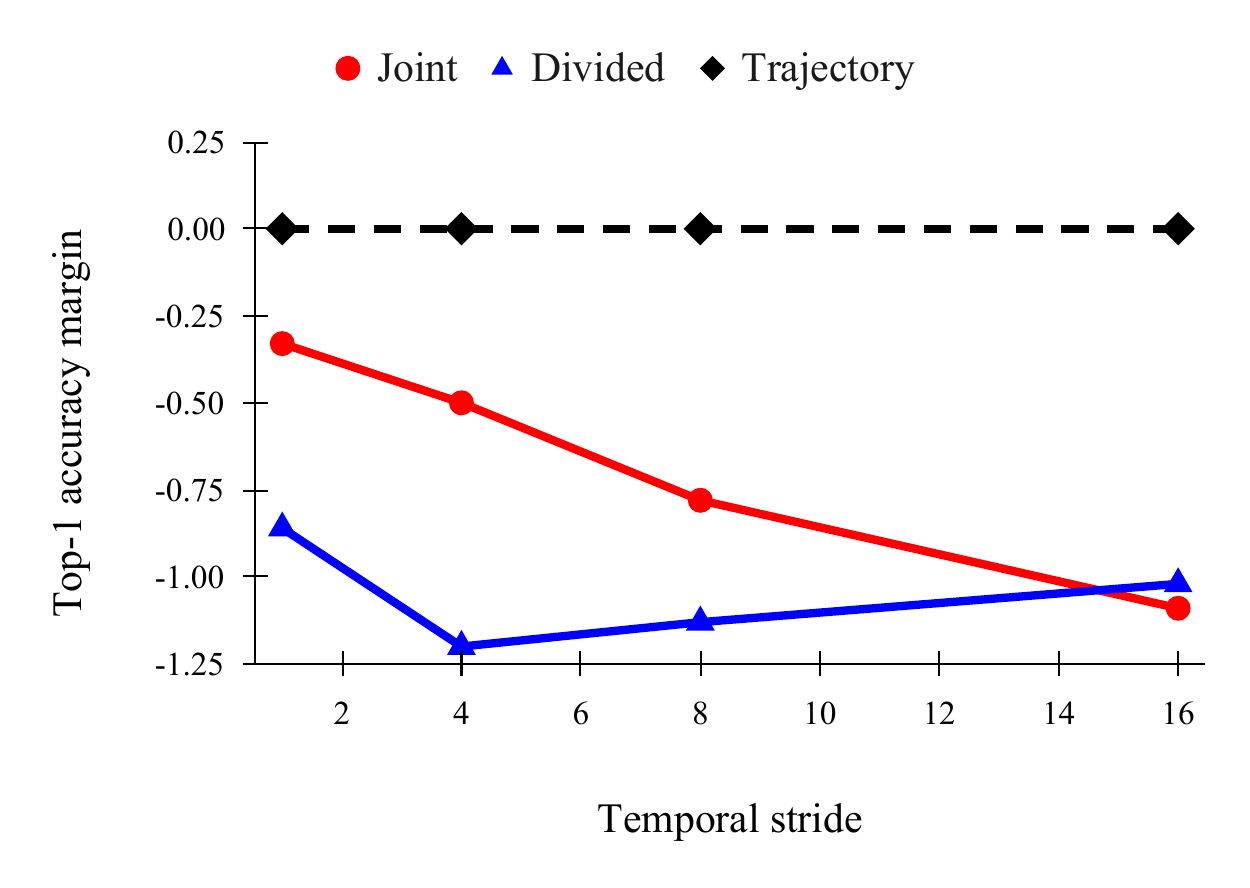}
        \subcaption{K-400: top-1 accuracy margin}
        \label{fig:motion_k400_rel}%
    \end{subfigure}\hfill%
    \begin{subfigure}[]{0.5\textwidth}\centering
    	\includegraphics[width=\textwidth]{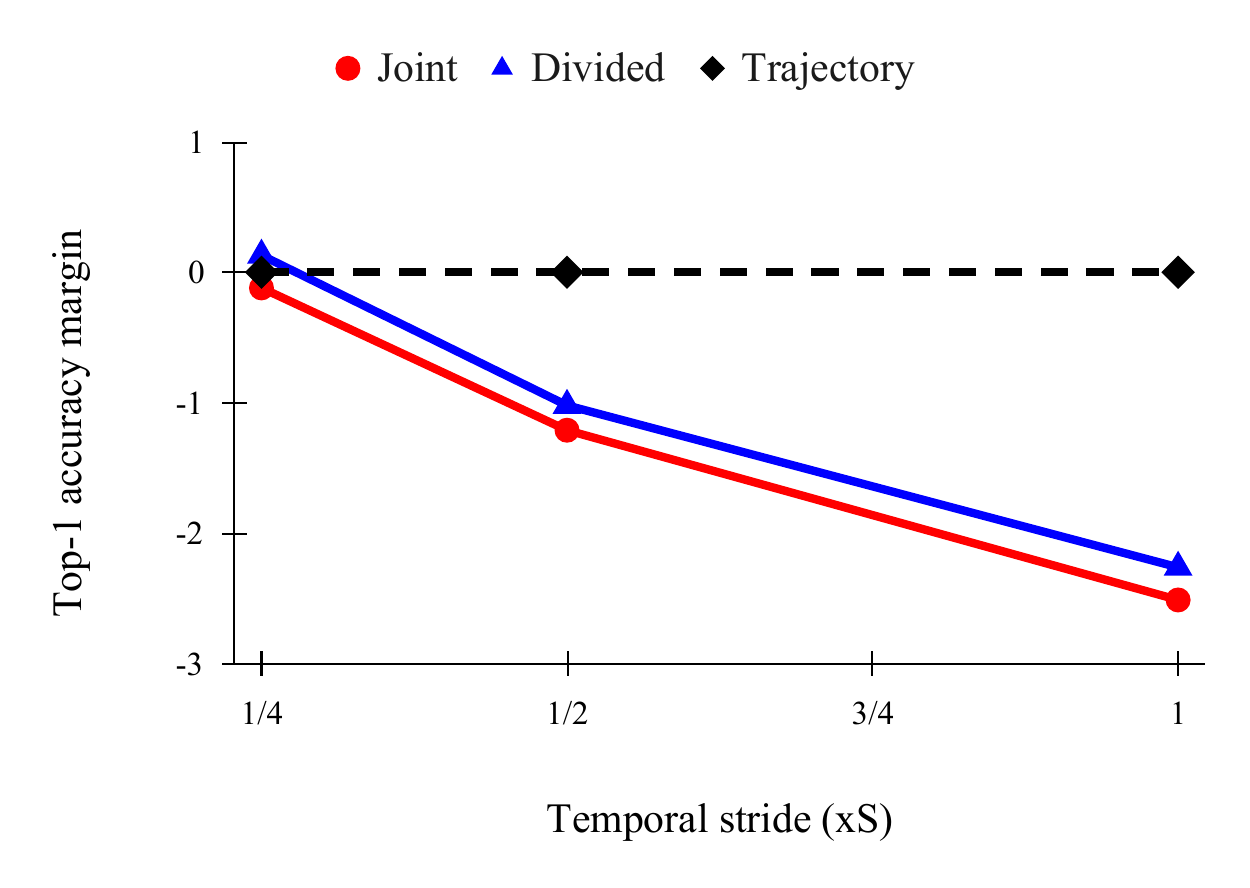}
    	\subcaption{SSv2: top-1 accuracy margin}
        \label{fig:motion_ssv2_rel}%
    \end{subfigure}\vfill%
    \begin{subfigure}[]{0.5\textwidth}\centering
    	\includegraphics[width=\textwidth]{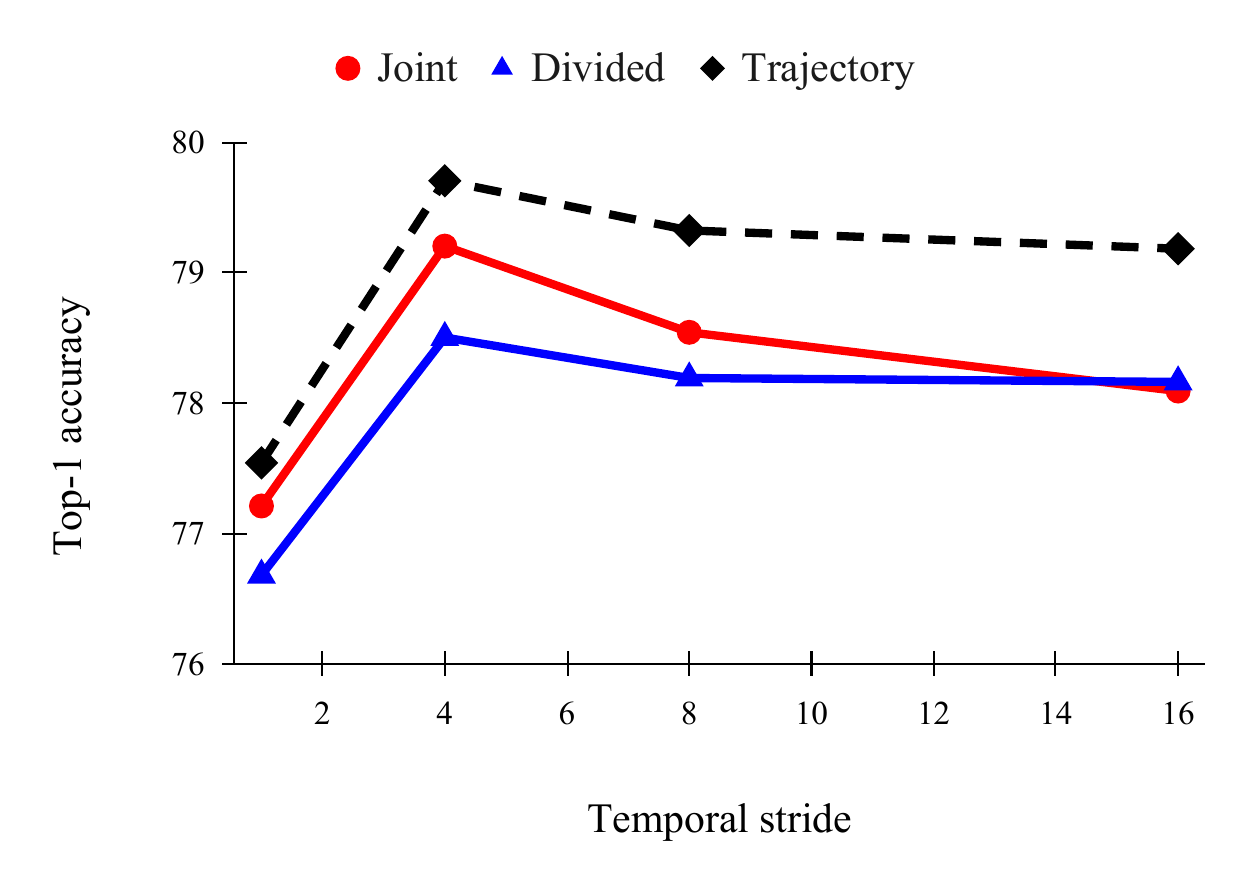}
        \subcaption{K-400: top-1 accuracy}
        \label{fig:motion_k400_rel}%
    \end{subfigure}\hfill%
    \begin{subfigure}[]{0.5\textwidth}\centering
    	\includegraphics[width=\textwidth]{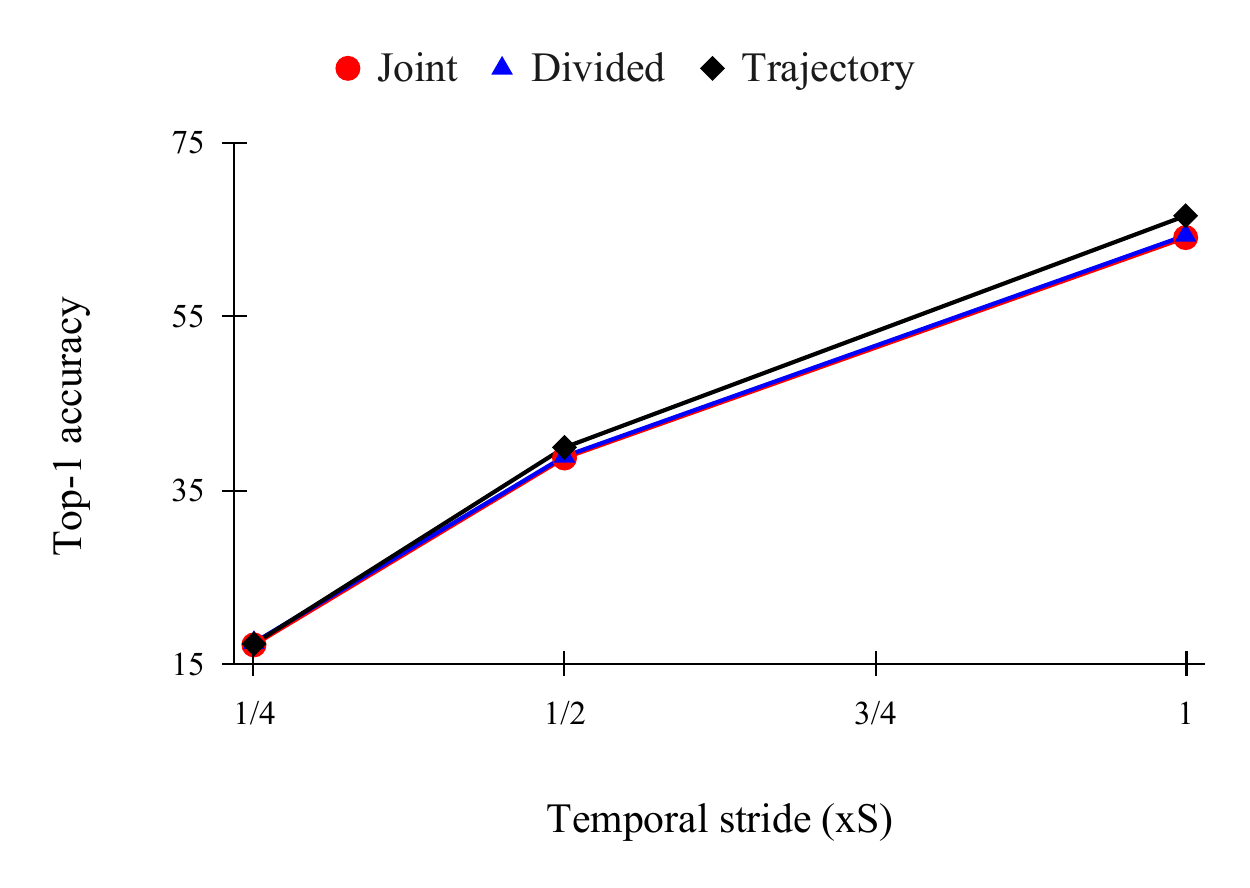}
        \subcaption{SSv2: top-1 accuracy}
        \label{fig:motion_ssv2_rel}%
    \end{subfigure}%
    \caption{
        \textbf{Does trajectory attention make better use of motion cues?}
        Performance of transformer models with joint space-time attention, divided space-time attention, and trajectory attention, as the temporal stride increases, on the Kinetics-400 dataset (left) and the Something--Something V2 dataset (right).
        Top: top-1 accuracy margin relative to trajectory attention (difference of accuracy and trajectory accuracy).
        Bottom: absolute top-1 accuracy shown for reference.
        If our trajectory attention is able to make better use of the temporal information in the video than the other attention mechanisms, we expect the accuracy margin between the methods to increase as the temporal stride increases. This is indeed the observed behaviour, especially for the motion cue-reliant SSv2 dataset.
        % The margin between our trajectory attention and the other methods increases as the temporal stride increases, especially for the motion cue-reliant SSv2 dataset.
        A larger stride simulates greater motion between input frames, which trajectory attention is better able to model and reason about.
        Note that the larger the stride, the more of the video clip is seen by the model; for all plots, the rightmost side of the axis corresponds to the entire video clip.
        Note also that the strides for SSv2 are written as multiples of $S$, the stride needed to evenly sample the entire video clip.
        % inference only, no training
    }\label{fig:motion}
\end{figure}

In the main paper (and below in \secref{s:datasets}), we provide evidence that action classification on the Something--Something V2 (SSv2) dataset~\cite{Goyal_2017} is more reliant on motion cues than the Kinetics dataset~\cite{kinetics}, where appearance cues dominate and a single-frame model achieves high accuracy. 
% single frame model: decrease of $39\%$ top-1 accuracy (SSv2) vs $7\%$ (K-400)
Improved performance on SSv2 is one way to infer that our model makes better use of temporal information, however, here we consider another way.
We artificially adjust the speed of the video clips by changing the temporal stride of the input.
A larger stride simulates faster motions, with adjacent frames being more different.
If our trajectory attention is able to make better use of the temporal information in the video than the other attention mechanisms, we expect the margin of improvement to increase as the temporal stride increases.
As shown in \figref{fig:motion}, this is indeed what we observe, with the lines diverging as temporal stride increases, especially for the motion cue-reliant SSv2 dataset.
Since the same number of frames are used as input in all cases, the larger the stride, the more of the video clip is seen by the model.
This provides additional confirmation that seeing a small part of a Kinetics video is usually enough to classify it accurately, as shown on the bottom left, where the absolute accuracy is reported.

\subsubsection{How important are motion cues for classifying videos from the Kinetics-400 and Something--Something V2 datasets?}\label{s:datasets}

% In the main paper , we provide evidence that action classification on the Something--Something V2 (SSv2) dataset~\cite{Goyal_2017} is more reliant on motion cues than the Kinetics dataset~\cite{kinetics}, where appearance cues dominate and a single-frame model achieves high accuracy. 
% single frame model: decrease of $39\%$ top-1 accuracy (SSv2) vs $7\%$ (K-400)

To determine the relative importance of motion cues compared to appearance cues for classifying videos on two of the major video action recognition datasets (Kinetics-400 and Something--Something V2), we trained a single frame vision transformer model and compare the results to a multi-frame model that can reason about motion.
The single frame was sampled from the video at random.
\tabref{abl_table:ssv2_k400} shows that single-frame action classifiers can do almost as well as video action classifiers on the Kinetics-400 dataset, implying that the motion information is much less relevant.
In contrast, classifying videos from the Something-Something V2 dataset clearly requires this motion information.
Therefore, to excel on the SSv2 dataset, a model must reason about motion information.
Our model, which introduces an inductive bias that favors pooling along motion trajectories, is able to do this and sees corresponding performance gains.

\begin{table*}[h]%\footnotesize
\centering
\caption{\textbf{Importance of motion cues for the K-400 and SSv2 datasets.} A classifier for the K-400 dataset performs well when all motion information is removed (1 frame model), while a classifier for the SSv2 dataset performs very poorly. Therefore, SSv2 is a better dataset for evaluating \textit{video} action classification, where the combination of appearance and motion is critical.}
\label{tab:ssv2_k400}
\setlength{\tabcolsep}{3pt}
% \begin{tabular}{l c | c c c}
% 	\toprule
% 	Dataset  & 1-frame RGB & 8-frames RGB & 8-frames flow & $\Delta$\\
% 	\midrule
% 	K-400          &     73.2         &    \bf 79.7          &     40.0          & 39.7  \\
% 	SSv2           &     27.1         &    \bf 66.5          &   47.0             & 19.5 \\
% 	 \bottomrule
% \end{tabular}

\begin{tabularx}{\linewidth}{l c c C}
	\toprule
	Dataset  & Top-1 accuracy (1 frame) & Top-1 accuracy (8 frames) & $\Delta$\\
	\midrule
	Kinetics-400          &     73.2  &    79.7 & 6.5  \\
	Something--Something V2           &     27.1  &    66.5 & \textbf{39.4} \\
	 \bottomrule
	\label{abl_table:ssv2_k400}
\end{tabularx}
\end{table*}

\subsubsection{Which classes is the performance difference larger with and without the trajectory attention?}\label{s:class_specific}
The class labels with the largest performance increase (given in parentheses) on the Something-Something v2 dataset are: “Spilling [something] next to [something]” (18\%), “Pretending to put [something] underneath [something]” (15\%), and “Trying to pour [something] into [something], but missing so it spills next to it” (14\%). The classes with the largest performance decrease are: “Putting [something] that can't roll onto a slanted surface, so it stays where it is” (10\%), “Putting [something] on a flat surface without letting it roll” (9\%), and “Showing a photo of [something] to the camera” (8\%). It is apparent that classes involving predominantly stationary objects do not benefit from trajectory attention, as we would expect.

\subsubsection{Trajectory attention maps}

\begin{figure}[t]
    \centering
    \includegraphics[ width=\linewidth]{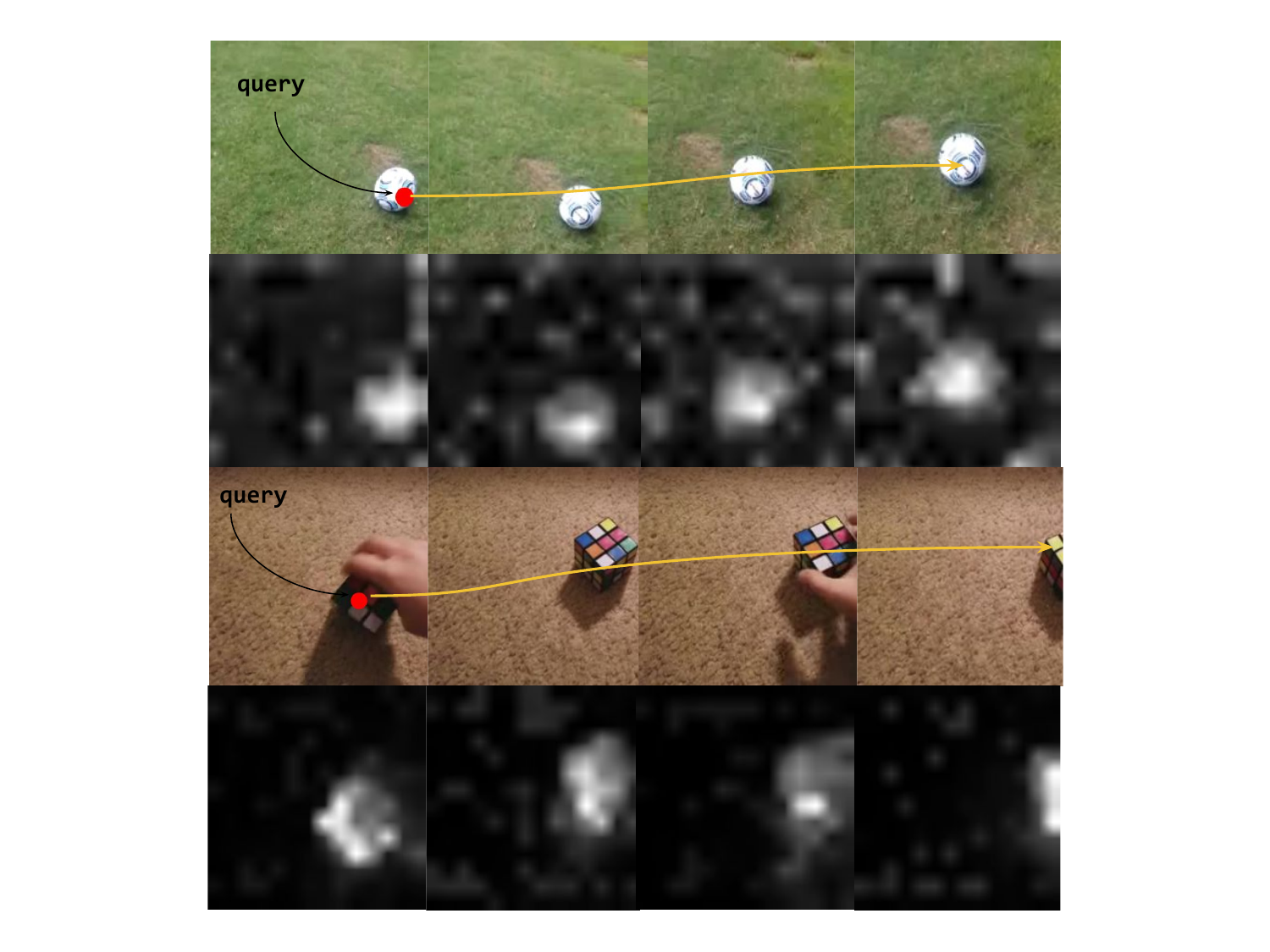}
    \caption{
        \textbf{Trajectory attention maps.}
        In this sequence of frames from Kinetics-400 (row 1) and Something-Something V2 (row 3), we show the attention maps at each frame given an initial query point (red point). We see that the model learns to implicitly track along motion paths (yellow arrow) using our trajectory attention module. 
    }\label{fig:attn_map}
\end{figure}

In \figref{fig:attn_map}, we show qualitative results of the intermediate attention maps of our trajectory attention operation. 
The learned attention maps appear to implicitly track the query points across time, a strategy that is easier to learn with the inductive bias instilled by trajectory attention.

% !TeX root = ../0_supplementary_material.tex

\subsubsection{How long does it take to train Motionformer model?}
For Table 3c, using Motionformer with orthoformer approximation, 16 prototypes take 384 GPU hours, 64 prototypes take 800 GPU hours, and 128 prototypes take 1216 GPU hours. 
For the Kinetics-400 state-of-the-art table, the Mformer-B model took 384 GPU hours, the Mformer-L took 1334 GPU hours, and Mformer-HR model took 1376 GPU hours to train. 
Our baseline Mformer-B model, which outperforms TimeSformer-B by over $1\%$, takes similar GPU hours (416 (ours) vs the 416 GPU hours reported in Table 2 of TimeSformer). 
We cannot directly compare to ViViT because they didn't report training time, but they used a very large transformer (24 layers) compared to ours (12 layers) and so we expect the training time for their approach to be significantly greater.

\subsubsection{Semi-supervised Video Object Segmentation on DAVIS 2017}
We evaluate our baseline Motionformer Kinetics-pretrained model (16x16 with Trajectory Attention) on the semi-supervised video object segmentation task on the DAVIS 2017 dataset as in Jabri et al.~\cite{jabri_space_time} in \tabref{tab:davis}. 
We directly use the attention maps of our Motionformer model in the label propagation setting, as in~\cite{jabri_space_time}. 
We report mean (m) of standard boundary alignment (F) and region similarity (J) metrics. 
We attain a competitive J\&F-Mean of 60.6. 
For comparison, DINO~\cite{caron2021emerging} obtains J\&F-Mean of 62.3 with the same architecture (ViT-B/16x16), but by using a self-supervised learning task on IM-1K. 
We expect that we could significantly improve the performance by using an 8x8 patch size, as this was shown to be highly effective for the task~\cite{caron2021emerging}.

% \begin{table}[t]%\footnotesize
% 	\centering
% 	\caption{\textbf{Can we train larger models using approximated trajectory attention?}
% 	We report top-1 and top-5 accuracy (\%) on the Kinetics-400 dataset of two variants of our Motionformer model: Motionformer-B and Motionformer-H.
% 	The former uses the base model with exact (E) trajectory attention, while the latter uses a much larger model (ViT-L) and a higher resolution input ($336\times336$ pixels) with approximate (A) trajectory attention, \ie, using Orthoformer.
% % 	The larger model has better performance, despite no optimization of the training schedule, hyperparameters, and data augmentation schedule.
% % 	The larger model also has far more parameters than the base model, and so unavoidably requires more GPU memory.
% % 	Furthermore, for a fixed patch size (in pixels), the memory requirements for exact attention scale with the square of the input resolution.
% 	We reduce this to a linear relationship with the Orthoformer approximation, which allows us to fit the model on the GPU.
% 	}
% 	\begin{tabularx}{\textwidth}{l c c c c C C}
% 	\toprule
% 		Model & Base model & Params & Attention & Max memory (GB) & Top-1 & Top-5 \\ 
% 		\midrule
% 		Mformer-B &  ViT-B/224 & 109.1M & Trajectory (E) & \textbf{7.3} & 79.7 & 94.2 \\ 
%  		Mformer-H &  ViT-L/336 & 381.9M & Trajectory (A) & 22.2 & \textbf{80.0} & \textbf{94.5} \\ 
% 		\bottomrule
% 	\end{tabularx}
% 	\vspace{-3mm}
% 	\label{tab:davis}
% \end{table}

\begin{table}[h]
\centering
	\caption{\textbf{DAVIS 2017 Video object segmentation.}
We evaluate the quality of frozen features on video instance tracking.
We report mean region similarity $\mathcal{J}_m$ and mean contour-based accuracy $\mathcal{F}_m$.
% We compare with existing self-supervised methods and a supervised ViT-S/8 trained on ImageNet.
% Image resolution is 480p.
}
  \setlength{\tabcolsep}{6pt}
\small
\begin{tabular}{@{}lllccc@{}}
\toprule
Method & Data & Arch. & $ (\mathcal{J}$\&$\mathcal{F})_m$ & $\mathcal{J}_m$ & $\mathcal{F}_m$\\
\midrule
\multicolumn{4}{@{}l}{\textit{Supervised}}\\
ImageNet & INet & ViT-S/8 & 66.0 & 63.9 & 68.1\\
STM~\cite{oh2019video} & I/D/Y & RN50 & 81.8 & 79.2 & 84.3 \\
\midrule
Ours & K-400 & Mformer-B/16 & 60.6 & 58.3 & 62.9 \\
\midrule
\multicolumn{4}{@{}l}{\textit{Self-supervised}}\\
CT~\cite{wang2019learning_cvpr} & VLOG & RN50 & 48.7 & 46.4 & 50.0\\
MAST~\cite{lai2020mast} & YT-VOS & RN18 & 65.5 & 63.3 & 67.6 \\
STC~\cite{jabri_space_time} & Kinetics & RN18 & \bf{67.6} & \bf{64.8} & \bf{70.2} \\
DINO~\cite{caron2021emerging} & INet & ViT-B/16 & 62.3 & 60.7 & 63.9 \\
\bottomrule
\end{tabular}
\label{tab:davis}
\end{table}

\subsection{Implementation details}
\paragraph{Preprocessing.}\label{s:experiments:imp_details}
During training, we randomly sample clips of size $16$\x$224$\x$224$ at a rate of $1/4$ from $30$ FPS videos, thereby giving an effective temporal resolution of just over $2$ seconds. 
We normalize the inputs with mean and standard deviation 0.5, rescaling in the range $[-1, 1]$.
We use standard video augmentations such as random scale jittering, random horizontal flips and color jittering.
For smaller datasets such as Something--Something V2 and Epic-Kitchens, we additionally apply rand-augment~\cite{Cubuk2019RandAugmentPD}. % and mixup~\cite{zhang2018mixup}.
During testing, we uniformly sample 10 clips per video and apply a 3 crop evaluation~\cite{slowfast}. 

\paragraph{Training.}
For all datasets, we use the AdamW~\cite{loshchilov2018fixing} optimizer with weight decay $5\times10^{-2}$, a batch size per GPU of 4, label smoothing~\cite{Inceptionv3} with alpha $0.2$ and mixed precision training~\cite{micikevicius2018mixed}.
For Kinetics-400/600 and Something-Something V2, we train for 35 epochs, with an initial learning rate of $10^{-4}$, which we decay by 10 at epochs 20, 30. 
As Epic-Kitchens is a smaller dataset, we use a longer schedule and train for $50$ epochs with decay at $30$ and $40$.

\paragraph{Long Range Arena benchmark details.}
For the Long-Range Arena benchmark~\cite{tay2021long}, we used the training, validation, and testing code and parameters from the \href{https://github.com/mlpen/Nystromformer}{Nystr\"omformer Github repository}.
The Performer~\cite{choromanski2021rethinking} implementation was ported over to PyTorch from the \href{https://github.com/google-research/google-research/tree/master/performer}{official Github repo}, and the Nystr\"omformer~\cite{xiong2021nystromformer} implementation was used directly from its Github repository.

\paragraph{Computing resources.}
Ablation experiments were run on a GPU cluster using 4 nodes (32 GPUs) with an average training time of 12 hours. 
Experiments for comparing with state-of-the-art models used 8 nodes (64 GPUs), with an average training time of 7 hours.

\paragraph{Libraries.}
For our code implementation, we used the \texttt{timm}~\cite{rw2019timm} library for our base vision transformer implementation, and the \texttt{PySlowFast}~\cite{fan2020pyslowfast} library for training, data processing, and the evaluation pipeline.

\end{document}